\documentclass{article}

\usepackage{microtype}
\usepackage{graphicx}
\usepackage{subcaption}
\usepackage{booktabs} 
\usepackage{xspace}
\usepackage{makecell}
\usepackage{tabularx}
\usepackage{multirow}

\usepackage{hyperref}

\usepackage[preprint]{icml2026}

\usepackage{amsmath}
\usepackage{amssymb}
\usepackage{mathtools}
\usepackage{amsthm}

\usepackage[capitalize,noabbrev]{cleveref}

\theoremstyle{plain}

\theoremstyle{definition}

\theoremstyle{remark}

\usepackage{pifont}
\usepackage{xcolor}
\newcommand{\cmark}{\textcolor{green!80!black}{\ding{51}}}
\newcommand{\xmark}{\textcolor{red}{\ding{55}}}
\newcommand{\dataname}{\textit{\textbf{Fine-T2I}}\xspace}
\usepackage{tcolorbox}
\tcbuselibrary{skins, breakable}
\newtcolorbox{chatbox}[1]{
  enhanced,
  colback=gray!5,      
  colframe=black,      
  boxrule=0.5pt,       
  arc=0pt,             
  colbacktitle=black,  
  coltitle=white,      
  fonttitle=\bfseries, 
  title=#1,
  attach boxed title to top left={yshift=-2mm,xshift=0mm},
  boxed title style={boxrule=0pt,arc=0pt,enhanced}
}
\usepackage[textsize=tiny]{todonotes}

\icmltitlerunning{Fine-T2I: An Open, Large-Scale, and Diverse Dataset for High-Quality T2I Fine-Tuning}

\begin{document}

\twocolumn[
  \icmltitle{Fine-T2I: An Open, Large-Scale, and Diverse Dataset \\for High-Quality T2I Fine-Tuning}



  \icmlsetsymbol{equal}{*}

  \begin{icmlauthorlist}
    \icmlauthor{Xu Ma}{yyy}
    \icmlauthor{Yitian Zhang}{yyy}
    \icmlauthor{Qihua Dong}{yyy}
    \icmlauthor{Yun Fu}{yyy}
    \\
    \raisebox{-0.5em}{\includegraphics[height=1.5em]{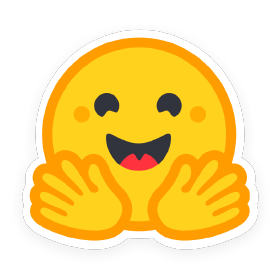}} Dataset: 
    \href{https://huggingface.co/datasets/ma-xu/fine-t2i}{{\small https://huggingface.co/datasets/ma-xu/fine-t2i}}
    \quad \raisebox{-0.5em}{\includegraphics[height=1.5em]{images/huggingface.png}}Space: 
    \href{https://huggingface.co/spaces/ma-xu/fine-t2i-explore}{{\small https://huggingface.co/spaces/ma-xu/fine-t2i-explore}}
  \end{icmlauthorlist}

  \icmlaffiliation{yyy}{Department of Electrical \& Computer Engineering, Northeastern University, Boston}

  \icmlcorrespondingauthor{Xu Ma}{ma.xu1@northeastern.edu}

  \icmlkeywords{Machine Learning, ICML}

  \vskip 0.3in
]



\printAffiliationsAndNotice{}  

\begin{abstract}

High-quality and open datasets remain a major bottleneck for text-to-image (T2I) fine-tuning. Despite rapid progress in model architectures and training pipelines, most publicly available fine-tuning datasets suffer from low resolution, poor text–image alignment, or limited diversity,  resulting in a clear performance gap between open research models and enterprise-grade models.
In this work, we present Fine-T2I, a large-scale, high-quality, and fully open dataset for T2I fine-tuning. 
Fine-T2I spans 10 task combinations, 32 prompt categories, 11 visual styles, and 5 prompt templates, and combines synthetic images generated by strong modern models with carefully curated real images from professional photographers. 
All samples are rigorously filtered for text–image alignment, visual fidelity, and prompt quality, with over 95\% of initial candidates removed. The final dataset contains over 6 million text–image pairs, around 2 TB on disk, approaching the scale of pretraining datasets while maintaining fine-tuning-level quality.
Across a diverse set of pretrained diffusion and autoregressive models, fine-tuning on Fine-T2I consistently improves both generation quality and instruction adherence, as validated by human evaluation, visual comparison, and automatic metrics. We release Fine-T2I under an open license to help close the data gap in T2I fine-tuning in the open community.
\end{abstract}

\section{Introduction}

\begin{table}[]
\centering
\setlength{\tabcolsep}{2pt} 
\resizebox{1\linewidth}{!}{
\begin{tabular}{lcccccc}
\toprule
\textbf{Dataset}     & \makecell{\textbf{High}\\\textbf{Resolution?}} & \makecell{\textbf{High}\\\textbf{Quality?}} & \makecell{\textbf{Designed}\\\textbf{Prompts?}} & \makecell{\textbf{Diverse}\\\textbf{Resolutions?}}  & \makecell{\textbf{Distribution}\\\textbf{Analysis?}} & \makecell{\textbf{Large}\\\textbf{Scale?}}\\
\midrule
\textbf{JourneyDB}&\cmark &\cmark &\xmark &\cmark &\xmark &\cmark\\
\textbf{Pick-a-Pic}&\xmark &\cmark &\xmark &\xmark &\xmark &\xmark\\
\textbf{T2I-2M} &\xmark &\xmark &\xmark &\xmark &\xmark &\cmark\\
\textbf{LAION-Art}&\xmark &\xmark &\xmark &\cmark &\xmark &\cmark\\
\textbf{LAION-Aesthetic}&\xmark &\xmark &\xmark &\cmark &\xmark &\cmark\\
\textbf{Blip3o-60k} &\xmark &\cmark &\xmark &\xmark &\xmark &\xmark\\
\textbf{Fine-T2I{\textcolor{cyan}{~\small(ours)}}} &\cmark &\cmark & \cmark& \cmark& \cmark &\cmark \\
\bottomrule
\end{tabular}
}
\vspace{1mm}
\caption{We compare our Fine-T2I with open-sourced fine-tuning datasets. High-resolution indicates the resolution of most samples is greater than 1K. More details can be found in Sec.~\ref{sec:Fine-T2I_dataset}. See Fig.~\ref{fig:compare_dataset} for a qualitative visual comparison illustrating the data quality.}
\label{tab:datasets_compare}
\vspace{-5mm}
\end{table}

\begin{figure}
    \centering
    \includegraphics[width=1\linewidth]{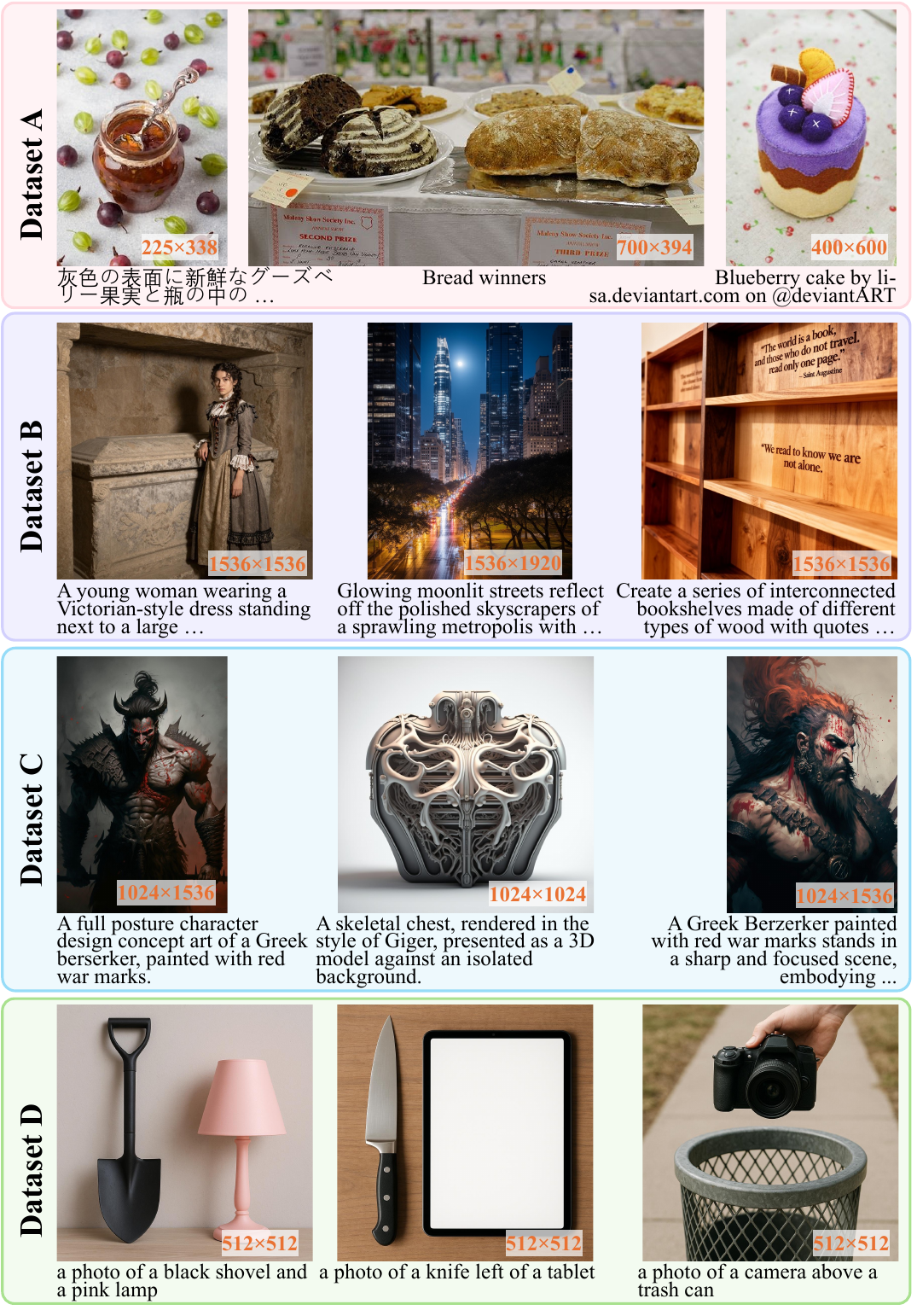}
    \vspace{-5mm}
    \caption{Visual comparison across datasets. For each dataset, we randomly sample three text–image pairs (no cherry-picking) to illustrate overall dataset quality. Zoom in for details. Dataset names for each row are provided on the following page.}
    \label{fig:guess_dataset}
\end{figure}

Text-to-image (T2I) generation has advanced rapidly in the most recent two years, enabling the synthesis of highly aesthetic images while faithfully adhering to diverse and complex instructions. This progress has been driven by a new wave of enterprise-grade generative models, including GPT Image 1.5, Nano Banana Pro~\cite{team2025gemma}, Seedream 4.0~\cite{seedream2025seedream}, and Qwen-Image~\cite{wu2025qwen}, among others. Such achievements always stem from two complementary factors: \textbf{model-centric} advances in architectures~\cite{wu2025qwen,gong2025seedream}, training objectives~\cite{liu2025flow,liu2025understanding}, and inference tricks~\cite{ma2025inference}; and \textbf{data-centric} progress in dataset scale, curation, and quality. Together, they push the frontier of high-fidelity image generation. 

While both model-centric advances and data-centric progress are critical in image generation, they do not propagate equally across our open community. 
Model-centric advances are always accessible. Model architectures,  training objectives, and inference techniques are routinely shared through publications or blogs and can be adopted by many teams. 
In contrast, the data-centric parts, especially carefully curated and instruction-aligned fine-tuning data, are often private and not accessible. 
As a result, SOTA models have become increasingly concentrated in top industry groups. Even when the open community matches the model training recipes and has enough computations, the lack of comparable high-quality data can lead to a persistent and potentially growing gap. As we can see, the text-to-image leaderboard~\footnote{https://huggingface.co/spaces/ArtificialAnalysis/Text-to-Image-Leaderboard} is dominated by enterprise-grade products.

\begin{figure*}[!t]
    \centering
    \includegraphics[width=1\linewidth]{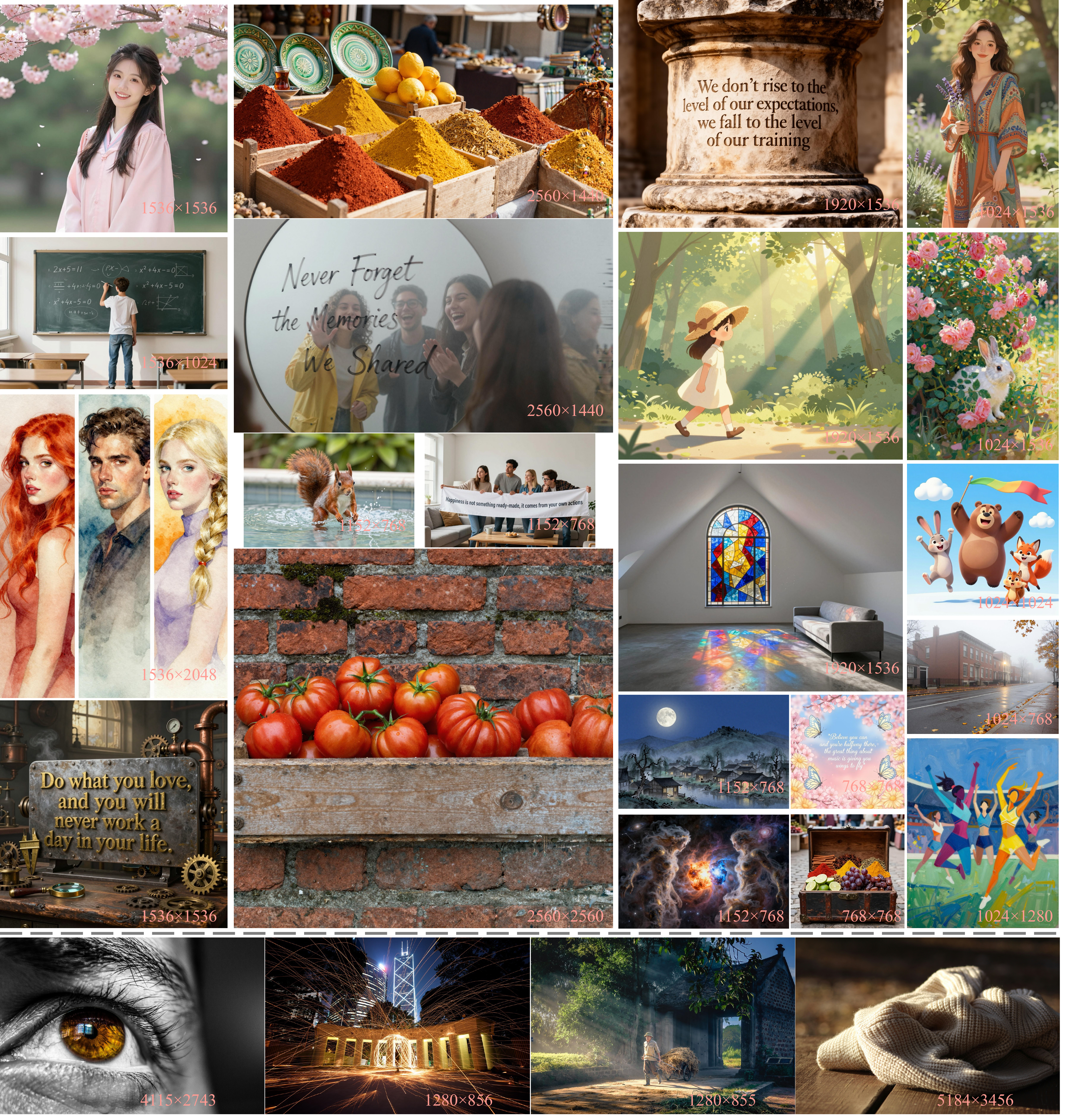}
    \vspace{-6mm}
    \caption{Examples of our introduced \dataname dataset samples, which include diverse resolutions, aspect ratios, styles, categories, tasks, \textit{etc}. Please check the supplementary for examples with detailed attributes and prompts. Images above the dashed line are our synthetic samples, and those below the dashed line are our curated real images. Please also refer our \href{https://huggingface.co/spaces/ma-xu/fine-t2i-explore}{Huggingface Space Page} to explore more.
    }
    \label{fig:example_images}
    \vspace{-3mm}
\end{figure*}

Similar to the open dissemination of model-centric techniques, the community has a strong interest in open-sourcing \emph{large-scale, high-quality} data for T2I training and alignment. Yet this is difficult for two reasons. First, truly high-quality fine-tuning images are extremely expensive (often \$10+ per image\footnote{https://stock.adobe.com/photos}). Second, such images are typically restricted by licenses and cannot be legally redistributed. Consequently, while many companies open-sourced the production-grade T2I models, high-quality fine-tuning datasets are rarely made public, and academic groups have no such resource to build a large-scale and high-quality fine-tuning dataset.  Therefore, the open community can only rely on datasets that are comparatively small, noisy, or distribution-specific, as summarized in Table~\ref{tab:datasets_compare}, which clearly bottleneck the training of practical strong models. 
Although these datasets established the groundwork for the field and made significant contributions, a clear gap remains in both scale and quality regarding the requirements for modern T2I fine-tuning.
We provide examples from each dataset in Fig.~\ref{fig:guess_dataset}, with dataset names presented in the footnote~\footnote{
\textbf{Dataset A:} \href{https://huggingface.co/datasets/laion/relaion-art}{LAION-art}; \ \ \ 
\textbf{Dataset B:} \href{https://huggingface.co/datasets/ma-xu/fine-t2i/tree/main/synthetic_original_prompt_random_resolution}{Our Fine-T2I (one synthetic subset)};  \ \ \ 
\textbf{Dataset C:} \href{https://huggingface.co/datasets/BLIP3o/BLIP3o-Pretrain-JourneyDB}{JourneyDB};  \ \ \ 
\textbf{Dataset D:} \href{https://huggingface.co/datasets/BLIP3o/BLIP3o-60k}{BLIP3o-60k (geneval\_train set).}}. Please see Fig.~\ref{fig:compare_dataset} for more examples.

To alleviate this bottleneck, we introduce Fine-T2I, a large-scale, high-quality, and fully open dataset for text-to-image fine-tuning. Fine-T2I is constructed to close the gap between (i) production-grade model releases and (ii) the lack of publicly available data that is both high-quality and legally licensed. Fine-T2I dataset mixes (a) synthetic data generated by strong diffusion models and (b) curated real images licensed and shared by photographers.
For the synthetic portion, we build the dataset from scratch, jointly designing prompts and generating images, covering diverse styles, categories, tasks, and prompt lengths, prompt formats, \textit{etc}. We further refine each prompt using a fine-tuned prompt enhancer model~\cite{wang2025promptenhancer}, and generate images at either a fixed square resolution or randomly sampled predefined aspect ratios, yielding four synthetic subsets.
For the curated real-image set, we use a strong VLM to generate both short and long prompts. 
We then design a detailed filtering pipeline for our data, including de-duplication, text–image alignment checks, safety guard, and aesthetic quality assessment, removing over 95\% of the initially collected data.
Finally, the resulting Fine-T2I contains roughly 6 million examples and occupies approximately 2 TB on-disk storage, offering exceptional scale and quality for open T2I fine-tuning. Fig.~\ref{fig:example_images} shows examples of our Fine-T2I dataset.
Unlike other open datasets that often assume fixed resolutions (\textit{e.g.}, $512\times512$), fixed prompt templates (\textit{e.g.}, ``a photo of \dots''), or simple settings (\textit{e.g.}, fixed object and position combinations), Fine-T2I targets a more practical goal: closing the gap between open models and production-grade generations. Fine-T2I is designed to improve human-preferred generation quality, aligning with large-scale human evaluation settings.

Beyond releasing the Fine-T2I dataset, we also provide a detailed pipeline for constructing our large-scale and high-quality dataset. Furthermore, we systematically validate its improvements across diverse T2I architectures, from autoregressive to diffusion models. 
With human evaluations, qualitative comparisons, and quantitative evaluations, we observe consistent and clear improvements when fine-tuning models on our Fine-T2I, indicating the effectiveness of our dataset. We hope that our Fine-T2I dataset, together with the full construction pipeline, can serve as a practical and beneficial foundation for image generation.

\section{Related Work}
\subsection{Text-to-Image Generation}
Currently, most text-to-image generation models are relying primarily on two architectural designs: autoregressive models, which predict discrete visual tokens sequentially~\cite{sun2024autoregressive,wang2024emu3,ma2025token,tian2024visual,li2024autoregressive}, and diffusion models~\cite{ho2020denoising,rombach2022high,lipman2022flow,dai2023emu,wu2025qwen}, which progressively remove noise from a randomly initialized Gaussian noise to construct an image.
Regardless of the differences in architectural designs, these models typically follow a progressive training strategy. First, we begin with large-scale pretraining on massive text-image pair datasets, like LAION-5B~\cite{schuhmann2022laion}, CC12M~\cite{changpinyo2021conceptual}, and OpenImages~\cite{kuznetsova2020open},  to learn broad world knowledge and semantic understanding. Next, models are fine-tuned on curated, high-aesthetic datasets, shifting the distribution to enhanced visual fidelity and detailed generation. To close the gap between original training objectives and human preference, recent methods incorporate Reinforcement Learning with Human Feedback (RLHF) such as Direct Preference Optimization (DPO)~\cite{rafailov2023direct,esser2024scaling,wang2024emu3} or Reinforcement Learning (RL)~\cite{liu2025flow,seedream2025seedream,wu2025rewarddance,wang2025ovis}. 
Crucially, this alignment is not a one-time process for SOTA image generation models. We see studies on a multi-turn training strategy. By iteratively training with human feedback alignment optimization and fine-tuning on high-quality datasets, the model progressively aligns with human preferences while mitigating reward hacking and maintaining training stability~\cite{deng2025openvlthinker}. That indicates the importance of SFT data in the training of a text-to-image generation system.

\subsection{Fine-tuning Datasets}
    
While fine-tuning is crucial for image generation quality, open and licensed high-quality datasets remain scarce, as discussed previously. Table~\ref{tab:datasets_compare} provides a detailed summarization for popular fine-tuning datasets, including JourneyDB~\cite{sun2023journeydb}, Pick-a-Pic~\cite{kirstain2023pick}, T2I-2M~\cite{jackyhate2024texttoimage2m}, LAION-Art~\cite{schuhmann2022laion}, LAION-Aesthetic~\cite{schuhmann2022laion}, and BLIP3o-60k~\cite{chen2025blip3}. While these datasets have laid the groundwork for open image generation research, they still exhibit clear gaps relative to modern production-grade models.
First, many samples from these datasets are low-resolution (often $\leq 1024\times1024$), whereas recent models (\textit{e.g.}, Qwen-Image~\cite{wu2025qwen}, SeedDream~\cite{gao2025seedream,seedream2025seedream}, and Nano Banana Pro~\cite{comanici2025gemini}) natively generate images at substantially higher resolutions. Second, even after applying aesthetics filters like LAION-Art and LAION-Aesthetic, the resulting data often falls short of the visual fidelity and aesthetics (see Fig.~\ref{fig:compare_dataset}) expected by today’s models. Most importantly, while always overlooked, existing open datasets rarely provide careful distribution analysis and organization, despite growing evidence that such curation is critical for strong generative performance~\cite{wu2025qwen,seedream2025seedream,cai2025z}. 
These limitations motivate the need for a new, high-quality fine-tuning dataset tailored to modern T2I generation.

\section{Fine-T2I Dataset}
\label{sec:Fine-T2I_dataset}
We introduce \textbf{Fine-T2I}, a new fine-tuning dataset designed for modern high-quality image generation in the open community. In this section, we describe the full pipeline, covering both the synthetic sets and the curated real-image set.
\subsection{Generating Synthetic Set}

The synthetic set constitutes the majority of Fine-T2I. We first construct controlled prompts with LLMs and apply rigorous filtering and de-duplication. Each retained prompt is then enhanced into a detailed, more descriptive counterpart using a fine-tuned prompt enhancer~\cite{wang2025promptenhancer}. Using both the original and enhanced prompts, we generate images with SOTA diffusion models~\cite{cai2025z,flux-2-2025} under two generation settings: (i) square aspect ratio and (ii) randomized resolutions. This yields four synthetic sub-datasets in total. Finally, we perform additional filtering to remove low-quality and poorly aligned samples, resulting in 6{,}145{,}693 high-quality synthetic samples.

\subsubsection{Prompt generation}

We generate synthetic prompts using the LLaMA3 instruction model~\cite{grattafiori2024llama}. 
To obtain a distribution that is both diverse and representative of real usage, we follow the dataset analyses reported in recent public reports (e.g., Qwen-Image~\cite{wu2025qwen} and Seedream~\cite{seedream2025seedream,gong2025seedream}) and complement them with trends reflected by the Image Arena leaderboard. 
Concretely, we design a \emph{weighted} taxonomy of high-level categories, including \textit{Nature}, \textit{Design}, \textit{People}, \textit{Text rendering}, and a set of \textit{rare cases}. 
We further decompose each category into fine-grained sub-categories to improve coverage and controllability.
We then synthesize \emph{controlled} prompts by composing the sampled category with several predefined attributes, including weighted style, prompt length, prompt structure (see supplementary Sec~\ref{sec:detailed_distribution_analysis}), and task type. 
Categories and styles are sampled according to calibrated weights to approximate real-world distribution. We emphasize frequently requested content like nature- and people-centric prompts and broadly useful styles like photorealistic, while still considering non-trivial probability mass to long-tail rare scenarios. 
The remaining attributes are sampled uniformly at random to increase variability.
To further reduce template outputs, we set a higher sampling temperature $1.4$ and a lower nucleus threshold top-$p=0.8$. 
We additionally filter out trivial generations by discarding prompts shorter than five words. 
In total, this yields 44{,}800{,}567 controlled prompts. 
We discuss practical issues encountered during prompt generation in Sec.~\ref{sec:problems}.

\begin{figure}
    \centering
    \includegraphics[width=1.0\linewidth]{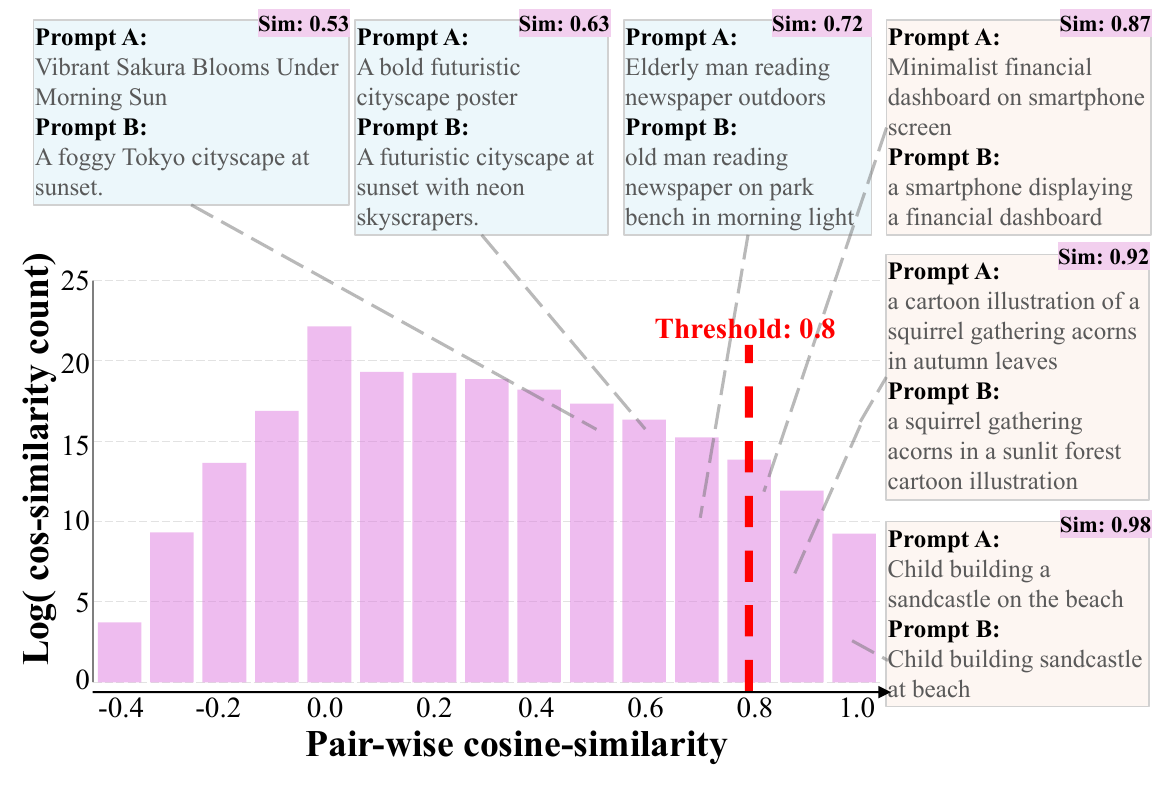}
    \vspace{-6mm}
    \caption{Semantic cosine-similarities distribution in a random prompt subset. We set deduplication threshold to 0.8.}
    \label{fig:deduplication}
     \vspace{-5mm}
\end{figure}
\subsubsection{Deduplication}
As a standard data-cleaning step~\cite{wu2025qwen,cai2025z,cao2025hunyuanimage}, de-duplication is required to remove identical or near-identical samples. 
For our synthetic prompt pool, this step is particularly important to increase diversity and to avoid over-representing repeated instructions. 
In practice, prompt de-duplication is commonly implemented either via (i) MinHash-LSH over token shingles (n-grams)~\cite{wu2025qwen} or (ii) embedding-based semantic filtering. 
Although MinHash-LSH is fast and scalable, shingle-based hashing is largely lexical-overlap driven. Prompts that share templates or phrasing may be removed even when their semantics differ. 
Since our prompts are LLM-generated and often vary in meaning with critical details under similar syntactic patterns, we adopt \emph{semantic} de-duplication.

We encode each prompt with the all-MiniLM-L6-v2 sentence encoder~\cite{reimers2019sentence,wang2020minilm} into a 384-dimensional vector embedding, and then we cache the embeddings for further processing. 
Computing full pairwise similarities is infeasible at our scale, so we employ a two-stage hierarchical strategy. 
We first partition the prompt set into 128 groups and perform de-duplication \emph{within} each group. Next, we run a second pass to remove remaining near-duplicates \emph{across} groups. In both stages, we treat two prompts as duplicates if their cosine similarity exceeds $0.8$, see Fig.~\ref{fig:deduplication} for illustration. After semantic de-duplication, we retain 5{,}555{,}147 diverse prompts.

\subsubsection{Prompt filtering}
\label{sec:prompt_filtering}
We next clean the generated prompt based on the content guard, alignment with corresponding attributes.

\paragraph{Guard}
To ensure the resulting prompts are safe and compliant, we apply content filtering with LLaMA-Guard-3-8B~\cite{inan2023llama}. 
We remove prompts flagged as \textit{violent crimes}, \textit{child sexual exploitation}, \textit{privacy}, and other unacceptable content. 
In addition, we filter by length and drop the prompts longer than 150 words.
Overall, this step provides 5{,}497{,}062 safety-checked prompts.

\paragraph{Filtering based on attributes}
In addition to the safety guard, we further filter prompts by checking whether each generated prompt aligns with its intended control attributes. 
Here, we focus on \emph{style} and \emph{category} consistency. 
We use Qwen3-VL-8B-Instruct~\cite{Qwen3-VL} as an attribute verifier. Given a prompt and its target attributes, the model judges whether the prompt semantically matches the specified style and category. 
Prompts that fail the alignment check are discarded. 
After attribute-alignment filtering, we obtain a final set of 5{,}158{,}969 prompts for subsequent image generation. See Sec.~\ref{sec:problems} for more discussions.

\subsubsection{Prompt Enhancing}
We see image generation models increasingly consider enhanced user inputs during inference. By rewriting the users' input into a more detailed instruction, models often provide better image quality and stronger alignment~\cite{ma2025token,wu2025qwen,comanici2025gemini}. We argue that this idea should not be viewed solely as an inference trick, it can also be valuable \emph{during training}. 
Motivated by this, we explicitly provide \emph{enhanced prompts} in Fine-T2I. For each filtered prompt, we apply a fine-tuned prompt enhancer~\cite{wang2025promptenhancer} to rewrite it into a longer, more descriptive counterpart while maintaining the original intent. The resulting enhanced prompts typically add additional visual attributes and reduce ambiguity. In our pipeline, we keep both counterparts, enabling models to benefit from diverse user input formats while also taking advantages from detailed instructions.

\subsubsection{Image generation}
We generate synthetic images from both prompt sets in Fine-T2I: the \emph{original} generated prompts and their \emph{enhanced} (rewritten) counterparts. For each variant, we generate images under two complementary settings. The first uses \emph{randomized resolutions and aspect ratios} to better reflect real-world generation requests and to encourage robustness beyond a single canonical format. Please check Tab.~\ref{tab:predefined_resolutions} for detailed resolutions.  The second restricts \emph{square} images with multiple square sizes, since many existing pipelines and benchmarks still primarily consider square inputs.

For each prompt, we prioritize quality via a generate-and-select strategy. We sample 1-3 candidate images using two strong open-source generators, Z-Image~\cite{cai2025z} and FLUX2~\cite{flux-2-2025}, and keep the candidate with the highest Aesthetic Predictor V2.5 score\footnote{https://github.com/discus0434/aesthetic-predictor-v2-5}. Compared with the commonly-used models like FLUX~\cite{labs2025flux1kontextflowmatching,flux2024}, GPT-4o, MidJourney used in T2I-2M, BLIP3-o, JourneyDB,  the aforementioned selected models can provide significantly better generation quality and stronger aesthetic appeal, which is critical for constructing a high-quality fine-tuning set. 
Combining the two prompt variants with the two resolution settings yields four synthetic subsets: \textit{Enhanced+Square}, \textit{Enhanced+Random}, \textit{Original+Square}, and \textit{Original+Random}. Since FLUX2 is substantially more expensive to run, most images in Fine-T2I are generated with Z-Image. Please check Fig.~\ref{fig:example_images} for generated examples.

\subsubsection{Text-Image pair Filtering}
\label{sec:text_image_pair_filtering}

Although we are using strong generators, the generated images are not always reliable. In practice, failures mainly arise from two reasons: (i) insufficient visual quality and aesthetics, and (ii) imperfect text-image alignment. We therefore apply a two-stage filtering pipeline for these issues.

We first filter images using Aesthetic Predictor V2.5 that focuses on image aesthetics. We keep an image only if its score $>5.5$, which can be considered very high visual quality. 
Filtering text-image alignment is substantially more challenging. Automated metrics like HPSv2~\cite{wu2023human} and HPSv3~\cite{ma2025hpsv3} provide useful insights, but we notice that they are insufficiently strict for fine-tuning data: they often miss fine-grained mismatches and fail to detect common generation artifacts. Indeed, we did not observe an off-the-shelf automatic metric that meets the high requirements needed for building a production-grade fine-tuning dataset.
To address this, we perform alignment verification (as well as artifacts check) with a VLM with reasoning ability or thinking mode. Although this brings much higher computational cost, we find that, under a carefully designed system prompt, the VLM becomes highly accurate at checking prompt compliance and typical failure modes in generated images. Our system prompt explicitly enumerates these alignment and artifact criteria. If a sample violates \emph{any} requirement, we filter it out. This reasoning-powered and VLM-based verification is crucial for ensuring that the remaining synthetic subset meets the quality bar required for fine-tuning. The detailed system prompt is provided in Fig.~\ref{fig:final_filter_system_prompt}. On average, we further filtered out about 70\% of text-image pairs in these steps, and the final samples are presented in Table~\ref{tab:final_synthetic_set}.

\begin{table}[!t]
\caption{Number of samples and storage for the synthetic set after filtering. ``PE" indicates enhanced prompts, ``PO" denotes original prompts, ``AR" stands for random aspect ratios, and ``AS" means square aspect ratio. }
\label{tab:final_synthetic_set}
\centering
\resizebox{0.96\linewidth}{!}{
\begin{tabular}{lcccc}
\toprule
\textbf{Subset}       & \textbf{PE-AR}    & \textbf{PE-AS}    & \textbf{PO-AR}    & \textbf{PO-AS}    \\
\midrule
\textbf{Final Samples}  & 1,615,592 & 1,538,253 & 1,686,498 & 1,305,350  \\
\textbf{Storage }      & 476G    & 517G    & 479G    & 436G     \\
\bottomrule
\end{tabular}
}
\vspace{-6mm}
\end{table}

\subsection{Curating Real-Image Set}
In addition to synthetic samples, Fine-T2I also includes a curated real-image set. Curating ``high-quality'' real data is inherently ambiguous: aesthetic quality is difficult to formalize, and purely automatic scorers can behave unpredictably on professional photography (\textit{e.g.}, subtle composition and lighting choices are not always captured by a single scalar score). Please see Sec.~\ref{sec:problems} for more discussions on image aesthetics and the current trend for human preference of generated images.
We therefore start from sources where aesthetic judgment is already exercised by humans. Specifically, we collect publicly available images from creator-driven platforms: \textit{Pexels}\footnote{https://www.pexels.com/}, \textit{Pixabay}\footnote{https://pixabay.com/}, and \textit{Unsplash-Lite}\footnote{https://unsplash.com/}, where photos are uploaded and curated by photographers and creators under open terms. This gives us a strong initial prior on aesthetics while retaining broad coverage of everyday photographic content.

We then impose additional quality constraints to meet the fine-tuning requirements. We filter images using Aesthetic Predictor V2.5 and keep only those with scores $\geq 6.5$, showing our stricter standard for this subset. We also discard extremely large images for potential training stability issue. 

\begin{table}[]
\caption{Number of samples for curated real-image set. We curated about 168k extremely high-quality samples from open-sourced platforms for this set in Fine-T2I.}
\label{tab:final_curated_set}
\resizebox{1\linewidth}{!}{
\begin{tabular}{lllll}
\toprule
\textbf{Data Source}    & \textbf{Downloaded }     & \textbf{After Filter} & \textbf{Kept ratio} & \textbf{Storage}   \\
\midrule
\textbf{Pexels}         & 233,342          & 117,389      & 50.3\%     & 192.6 GB \\
\textbf{Pixabay}        & 163,524          & 32,654       & 20.0\%     & 9.7 GB   \\
\textbf{Unsplash\_lite} & 24,997           & 18,381       & 73.6\%     & 56.2 GB \\
\bottomrule
\end{tabular}
}
\vspace{-5mm}
\end{table}

To make these images qualified for text-conditioned training, we generate captions with a fine-tuned\footnote{https://huggingface.co/Ertugrul/Qwen2.5-VL-7B-Captioner-Relaxed} Qwen2.5-VL-7B model~\cite{bai2025qwen2}. Same as the synthetic set, we provide two prompts for each image: an initial prompt (short length) and a more detailed enhanced version generated by the same prompt enhancement procedure. This mirrors the way prompts are used in practice, from short user inputs to more explicit, descriptive instructions, and provides richer instructions for fine-tuning. After filtering and captioning steps, the real-image subset contains 168{,}424 images. The detailed statistics are presented in Table~\ref{tab:final_curated_set}.

\section{Fine-T2I Dataset Specifications}

\begin{figure*}[!t]
    \centering
    \begin{subfigure}[b]{0.33\linewidth}
        \centering
        \includegraphics[width=\linewidth]{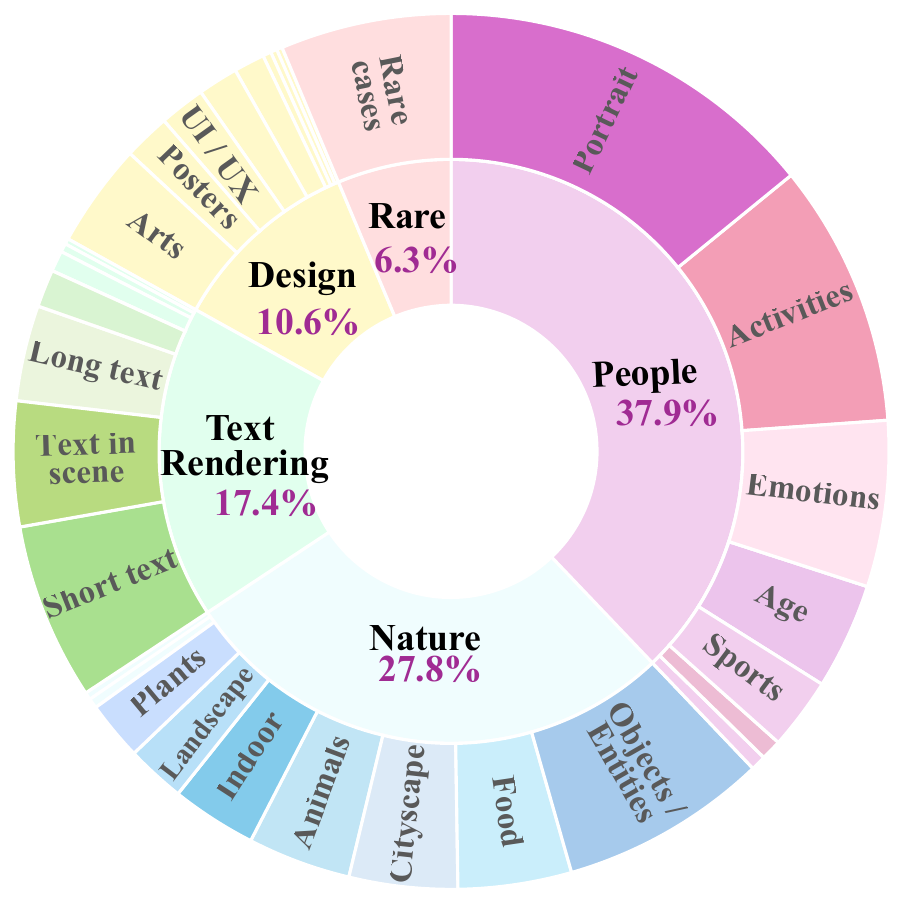}
        \vspace{-6mm}
        \caption{Categories Analysis} 
        \label{fig:category}
    \end{subfigure}
    \hfill 
    \begin{subfigure}[b]{0.33\linewidth}
        \centering
        \includegraphics[width=\linewidth]{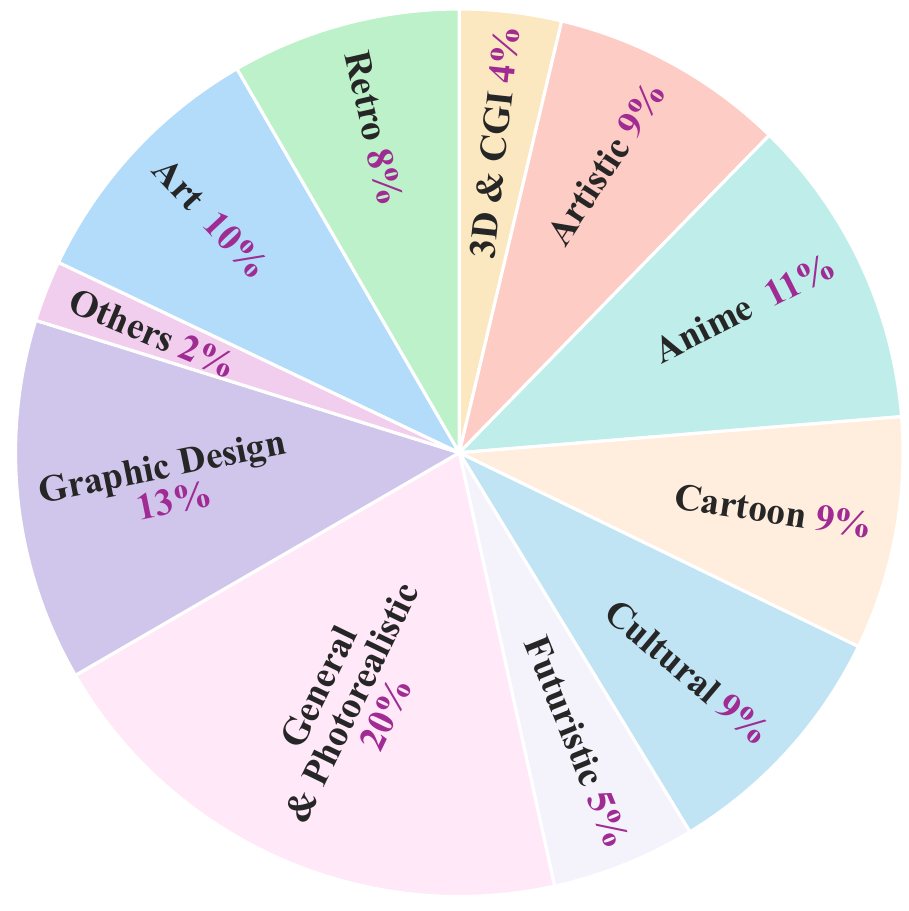}
        \vspace{-6mm}
        \caption{Styles Analysis} 
        \label{fig:style}
    \end{subfigure}
    \hfill 
    \begin{subfigure}[b]{0.33\linewidth}
        \centering
        \includegraphics[width=\linewidth]{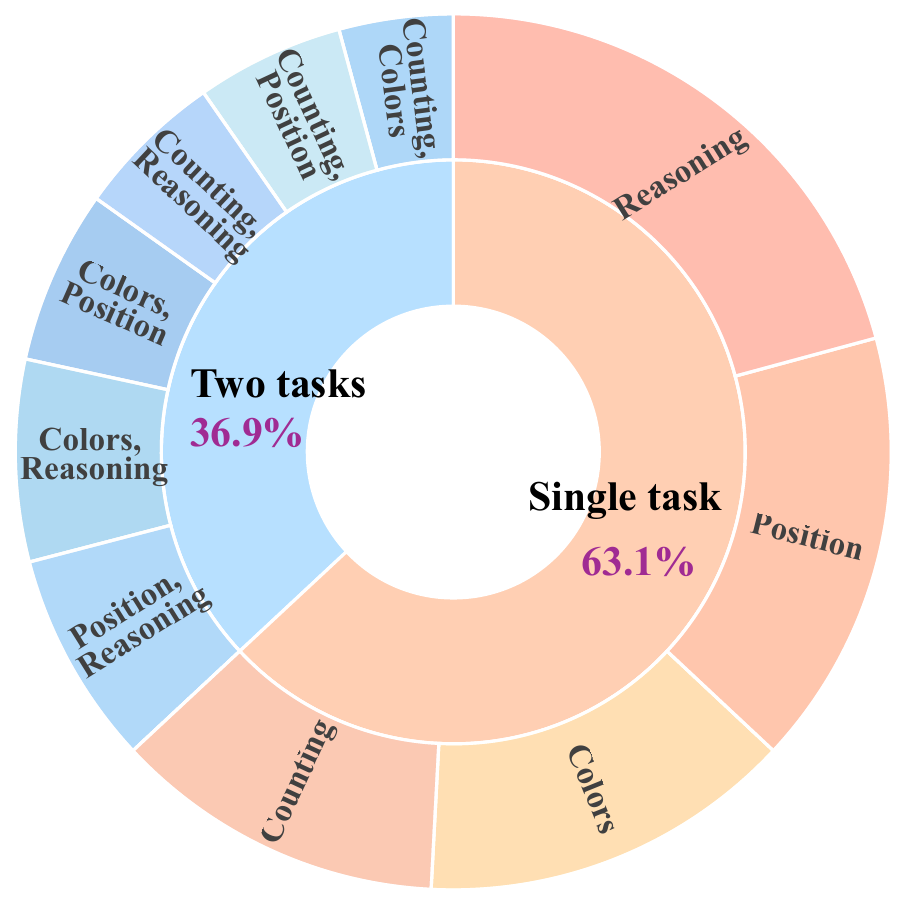}
        \vspace{-6mm}
        \caption{Tasks Analysis} 
        \label{fig:task}
    \end{subfigure}
    \vspace{-6mm}
    \caption{Analysis for our synthetic set, we provide the distribution for sample categories, sample styles, and the related tasks. Notice that for the tasks, we only consider the prompts that have specific requirements for the task when generating the prompts, while about 61.5\% of prompts did not ask for specific tasks. Please check Sec.~\ref{sec:detailed_distribution_analysis} in the supplementary for details.}
    \label{fig:analysis_3_attributes}
    \vspace{-3mm}
\end{figure*}

The above section detailed the pipeline for building our dataset. Next, we provide detailed insights into our dataset.

\vspace{-4mm}
\paragraph{Analysis on the attributes} We analyze the distribution of Fine-T2I to understand its coverage and balance. Since the synthetic set dominates the dataset and comes with explicit attribute labels, we focus our analysis on the synthetic subset. Fig.~\ref{fig:analysis_3_attributes} summarizes the distributions over \textbf{categories}, \textbf{styles}, and \textbf{tasks}. 
As shown in Fig.~\ref{fig:category}, the category mixture is focusing on realistic, high-demand content: \textit{People} (37.9\%) and \textit{Nature} (27.8\%) form the majority, while \textit{Text Rendering} (17.4\%) and \textit{Design} (10.6\%) provide substantial coverage of instruction-heavy and layout-sensitive cases, and \textit{Rare cases} (6.3\%) ensures non-trivial long-tail support. 
Fig.~\ref{fig:style} indicates that styles are similarly balanced: \textit{General \& Photorealistic} is the largest portion (20\%), complemented by diverse artistic and design-oriented styles, which helps the dataset span both photorealistic generation and stylized creation. 
Finally, Fig.~\ref{fig:task} shows that while most prompts correspond to \textit{single-task} instructions (63.1\%), a sizable fraction involves \textit{two-task} compositions (36.9\%), covering attributes such as reasoning, counting, colors, and positions. These tasks provide meaningful instructions on hard sample generation. Please also check Fig.~\ref{fig:prompt_length} in the supplementary for the study of prompt length distribution.

Although these empirical distributions deviate slightly from the initial sampling weights (due to non-uniform rejection during safety, alignment, and quality filtering), the final dataset still closely matches the intended real-world demand. It emphasizes common user scenarios without collapsing diversity, and preserves long-tail and multi-attribute coverage, which is extremely critical for training strong modern T2I models like Qwen-Image or Z-Image.

\begin{figure}[!t]
    \centering
    \vspace{-2mm}
    \includegraphics[width=1\linewidth]{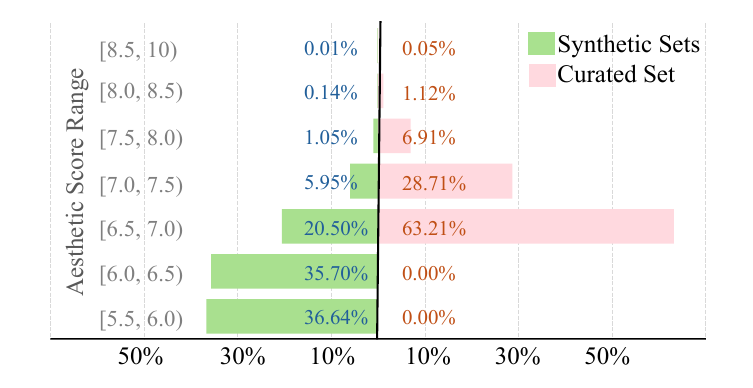}
    \vspace{-7mm}
    \caption{The aesthetic score distribution of our Fine-T2I. Both the synthetic sets and the curated set demonstrate high aesthetic scores, implying strong visual quality for fine-tuning.}
    \label{fig:aesthetic_distribution}
    \vspace{-3mm}
\vskip -0.1in
\end{figure}

\paragraph{Analysis on the aesthetic quality}
We further examine the aesthetic quality of Fine-T2I using the Aesthetic Predictor V2.5 scores, with distributions for the synthetic and curated real-image sets shown in Fig.~\ref{fig:aesthetic_distribution}. 
For this metric, scores above 5.5 are generally regarded as indicative of high aesthetic quality.
We have two observations from the distribution. First, the curated real-image subset is more concentrated at higher scores, with the majority falling in $[6.5,7.0)$ (63.21\%) and a substantial portion in $[7.0,7.5)$, plus a non-trivial tail beyond $7.5$. This is consistent with our sourcing strategy from creator-driven photography platforms and our stricter filtering criteria. 
Second, the \emph{synthetic subset} exhibits a broader score distribution. Most samples lie in the range of $[5.5,6.5)$ (36.64\% in $[5.5,6.0)$ and 35.70\% in $[6.0,6.5)$), while a meaningful fraction reaches higher aesthetic ranges, with a small tail above $7.5$. This reflects the inherent variability of generative models and our use of a permissive threshold to preserve scale and diversity. Overall, the two subsets are complementary. The curated real-image set provides a consistently high-aesthetic standard, while the synthetic set contributes large-scale, diverse guidance with controlled attributes.

\section{Experiments with Fine-T2I}
\begin{figure}[!t]
    \centering
    \begin{subfigure}[b]{0.95\linewidth}
        \centering
        \includegraphics[width=1.0\linewidth]{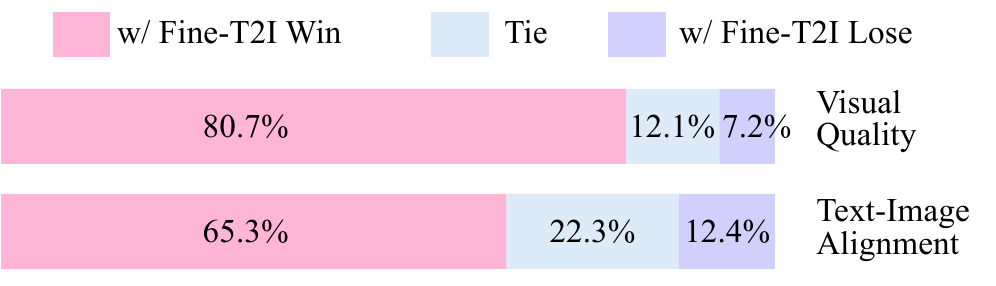}
        \vspace{-7mm}
        \caption{Human-evaluation results on LlamaGen.} 
        \label{fig:human_eval_llamagen}
        \vspace{3mm}
    \end{subfigure}
    
    \begin{subfigure}[b]{1\linewidth}
        \centering
        \vspace{-3mm}
        \includegraphics[width=0.95\linewidth]{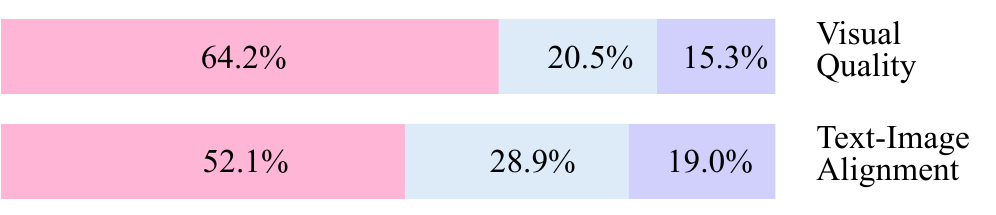}
        \vspace{-2mm}
        \caption{Human-evaluation results on SD-XL.} 
        \label{fig:human_eval_sdxl}
    \end{subfigure}
    \hfill 
    \caption{Human-evaluation results for the autoregressive model LlamaGen and the diffusion model SD-XL. We compare generations produced with and without fine-tuning on our dataset.}
    \label{fig:human_eval}
\vskip -0.15in
\end{figure}

\begin{figure}[!h]
    \centering
    \includegraphics[width=1\linewidth]{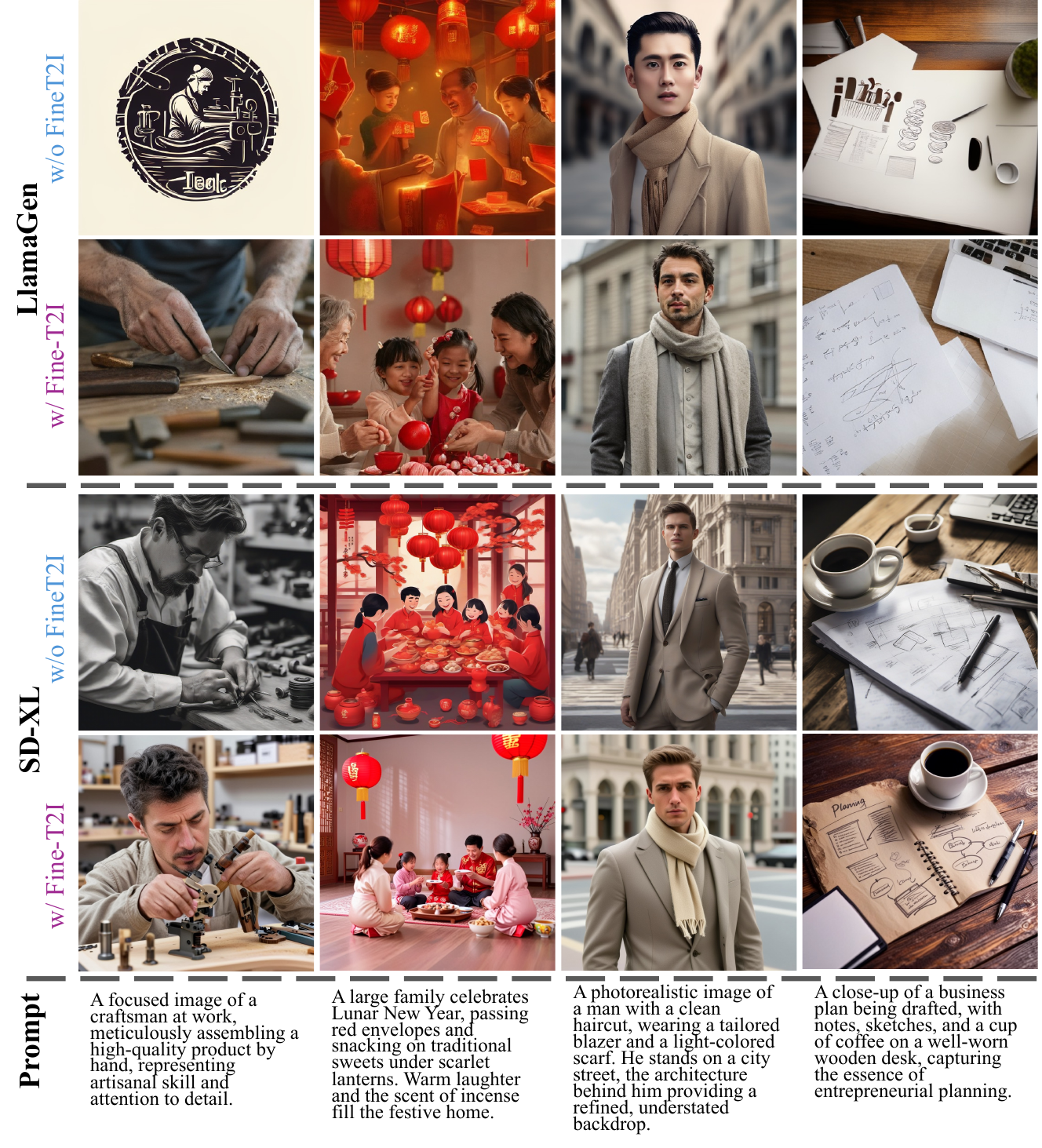}
    \caption{Visual comparison between original generations and generations fine-tuned on our Fine-T2I. One can observe clearly better generation quality on our fine-tuned model.}
    \label{fig:finetune_visual_compare}
\vskip -0.05in
\end{figure}

We next evaluate Fine-T2I for image generation, investigating two problems: whether Fine-T2I consistently improves generation quality and instruction following across model families, and (ii) which aspects Fine-T2I improves the most.

\paragraph{Experimental Setting}
Rather than focusing on a single modeling family, we consider two representative T2I backbones spanning different generation technologies: a diffusion-based model SD-XL~\cite{podell2023sdxl} and an autoregressive model LlamaGen~\cite{sun2024autoregressive}.
We select these two models because they are not well-trained on closed high-quality datasets like Qwen-Image or Nano-Banana, better demonstrating the performance gap.
For each model, we start from the official publicly released pretrained checkpoint and perform continued fine-tuning on Fine-T2I. Fine-T2I includes four synthetic subsets (original/enhanced $\times$ square/random resolution) as well as a curated real-image set; unless otherwise specified, we fine-tune the two models using a mixture of all subsets.
For SD-XL, we fine-tune with LoRA adapters to ensure a lightweight training setup, a batch size of 8 on each GPU and a learning rate of $1\times10^{-4}$. For LlamaGen, we fine-tune the full model using batch size 24 and learning rate $3\times10^{-5}$. 
We filter out samples with relatively long prompts, considering the limited token length supported by the two models. 
We fine-tune each model with around 1 epoch to avoid overfitting our dataset.  While better settings may further improve the fine-tuning results, that is not our main target here.

\paragraph{Evaluation}
As noted in HunyuanImage~\cite{cao2025hunyuanimage} and related work, commonly used automatic benchmarks such as GenEval~\cite{ghosh2023geneval} and T2I-CompBench~\cite{huang2023t2i} have limited coverage and can be misaligned with human preference, due to restricted prompt diversity. This limitation is especially pronounced when the goal is to assess the quality of a fine-tuning dataset, where improvements may manifest in subtle aspects such as aesthetics, style fidelity, and nuanced instruction following. 

To better reflect real-world usage and human judgment, we therefore construct an evaluation suite by randomly sampling 500 public prompts from the Artificial Analysis Image Arena leaderboard. These prompts cover diverse user requests and have been widely used to compare the leading T2I models. We conduct large-scale human preference evaluation on the resulting generations, focusing on text-image alignment and overall human-preferred visual quality. For completeness, we also report GenEval results in Table~\ref{tab:geneval}, which provide a reference point for improvements on established automatic protocols. But we emphasize that we should not heavily rely on these automated metrics.

\begin{table}[]
\caption{GenEval benchmark results for LlamaGen and SD-XL. While this auto-evaluation does not fully capture the benefits of our dataset, consistent improvements are still observed.}
\label{tab:geneval}
\setlength{\tabcolsep}{3pt} 
\resizebox{1\linewidth}{!}{
\begin{tabular}{llllllll}
\toprule
\textbf{Model}      & \textbf{Single} & \textbf{Two}  & \textbf{Count} & \textbf{Color} & \textbf{Position} & \textbf{Attribute} & \textbf{Overall} \\
\textbf{LlamaGen}   & 0.71   & 0.34 & 0.21     & 0.58   & 0.07     & 0.04      & 0.32    \\
{\footnotesize \textbf{\textcolor{cyan}{ - w/ Fine-T2I}}} &  0.82      &  0.40    &  0.37      &  0.74      &   0.11       &   0.14        &   0.41  {\scriptsize \textbf{\textcolor{cyan}{ (+0.09)}}}    \\
\midrule
\textbf{SD-XL}      & 0.98   & 0.74 & 0.39     & 0.85   & 0.15     & 0.23      & 0.55    \\
{\footnotesize \textbf{\textcolor{cyan}{- w/ Fine-T2I}}} &   0.98     &  0.79   &    0.40      &   0.89     &   0.21       &   0.27        &  0.61{\scriptsize \textbf{\textcolor{cyan}{~(+0.06)}}}     \\
\bottomrule
\end{tabular}
}
\vskip -0.15in
\end{table}

\paragraph{Results analysis}
We primarily evaluate the improvements from fine-tuning on Fine-T2I based on human evaluation, since our goal is to assess Fine-T2I as a \emph{fine-tuning dataset} and the improvements we target (overall aesthetics, realism, and instruction following) are not reliably captured by automatic benchmarks. As shown in Fig.~\ref{fig:human_eval},
fine-tuning on Fine-T2I leads to clear preference gains on both models. By fine-tuning LlamaGen on our dataset, we achieve an 80.7\% win rate for visual quality compared to a counterpart without further fine-tuning, and a 65.3\% win rate for text-image alignment. We also see substantive improvements for SD-XL. These observations suggest that our Fine-T2I is able to boost different architectures and generation designs, providing consistent benefits across different model families. Meanwhile, we also notice that our dataset boosts the visual quality the most. This might be due to the extremely high aesthetic quality of our dataset.

The qualitative examples are presented in Fig.~\ref{fig:finetune_visual_compare}. After fine-tuning on  Fine-T2I, generations look more natural and coherent, with cleaner local details and fewer distracting artifacts, while better reflecting the intended prompt content. Although we do not view it as the main evidence, GenEval also improves for both models, as shown in Table~\ref{tab:geneval}, providing an additional signal that Fine-T2I enhances controllability on standard automatic protocols.

\begin{figure}[!t]
    \centering
    \includegraphics[width=1\linewidth]{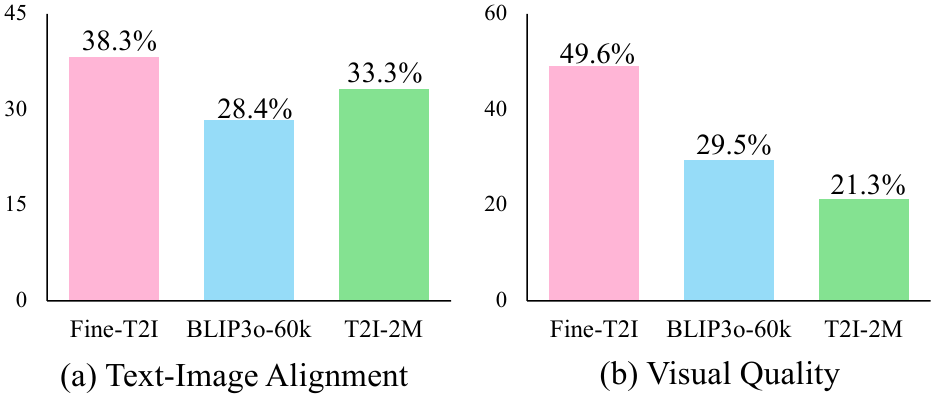}
    \vspace{-5mm}
    \caption{Human evaluation results (win rate) comparing models fine-tuned on different datasets. For each comparison, the best generation result among the three is selected as the winning example.}
    \label{fig:human_compare_data}
    \vspace{-5mm}
\end{figure}

\paragraph{Comparing with other SFT datasets}
Previous studies demonstrate the high-quality of our Fine-T2I and the benefits of our dataset. Here, we compare our dataset with other datasets. We use LlamaGen as the candidate model to finetune considering the simplicity. We finetune LlamaGen with the same settings on T2I-2M, BLIP3o-60k, and our Fine-T2I, respectively. We do a human evaluation (select the best from three generations) on the 500 prompts as mentioned before, and report the win rate results in Fig.~\ref{fig:human_compare_data}. The generated images can be found in Fig.~\ref{fig:compare_4} for visual comparison.  Clearly, models fine-tuned on Fine-T2I dataset generate much better results than the other two datasets on both text alignment and visual quality. We also suggest referring to Fig.~\ref{fig:compare_dataset} for visual quality comparison of the samples in different fine-tuning datasets.

\section{Conclusion}
Instead of model-centric innovations, public and open high-quality fine-tuning data remains a key bottleneck for high-quality text-to-image generation in the open community. 
To address this gap, we introduce Fine-T2I, a large-scale, fully open dataset (over 6M samples) that combines diverse synthetic text–image pairs with a rigorously curated real-image set. 
Fine-T2I is built with a systematic data design and filtering pipeline that prioritizes diversity, alignment, and visual quality at scale.
We show that fine-tuning on Fine-T2I consistently improves different model families, yielding stronger instruction adherence and much higher visual quality. 
With Fine-T2I released under an open license, we aim to provide a solid foundation for future work on text-to-image generation, and pave the way to promising modern text-to-image models in the open community.



\nocite{langley00}

\bibliography{paper}

@article{team2025gemma,
  title={Gemma 3 technical report},
  author={Team, Gemma and Kamath, Aishwarya and Ferret, Johan and Pathak, Shreya and Vieillard, Nino and Merhej, Ramona and Perrin, Sarah and Matejovicova, Tatiana and Ram{\'e}, Alexandre and Rivi{\`e}re, Morgane and others},
  journal={arXiv preprint arXiv:2503.19786},
  year={2025}
}

@article{lipman2022flow,
  title={Flow matching for generative modeling},
  author={Lipman, Yaron and Chen, Ricky TQ and Ben-Hamu, Heli and Nickel, Maximilian and Le, Matt},
  journal={arXiv preprint arXiv:2210.02747},
  year={2022}
}

@article{wang2024emu3,
  title={Emu3: Next-token prediction is all you need},
  author={Wang, Xinlong and Zhang, Xiaosong and Luo, Zhengxiong and Sun, Quan and Cui, Yufeng and Wang, Jinsheng and Zhang, Fan and Wang, Yueze and Li, Zhen and Yu, Qiying and others},
  journal={arXiv preprint arXiv:2409.18869},
  year={2024}
}

@inproceedings{esser2024scaling,
  title={Scaling rectified flow transformers for high-resolution image synthesis},
  author={Esser, Patrick and Kulal, Sumith and Blattmann, Andreas and Entezari, Rahim and M{\"u}ller, Jonas and Saini, Harry and Levi, Yam and Lorenz, Dominik and Sauer, Axel and Boesel, Frederic and others},
  booktitle={Forty-first international conference on machine learning},
  year={2024}
}

@article{rafailov2023direct,
  title={Direct preference optimization: Your language model is secretly a reward model},
  author={Rafailov, Rafael and Sharma, Archit and Mitchell, Eric and Manning, Christopher D and Ermon, Stefano and Finn, Chelsea},
  journal={Advances in neural information processing systems},
  volume={36},
  pages={53728--53741},
  year={2023}
}

@article{dai2023emu,
  title={Emu: Enhancing image generation models using photogenic needles in a haystack},
  author={Dai, Xiaoliang and Hou, Ji and Ma, Chih-Yao and Tsai, Sam and Wang, Jialiang and Wang, Rui and Zhang, Peizhao and Vandenhende, Simon and Wang, Xiaofang and Dubey, Abhimanyu and others},
  journal={arXiv preprint arXiv:2309.15807},
  year={2023}
}

@article{ma2025inference,
  title={Inference-time scaling for diffusion models beyond scaling denoising steps},
  author={Ma, Nanye and Tong, Shangyuan and Jia, Haolin and Hu, Hexiang and Su, Yu-Chuan and Zhang, Mingda and Yang, Xuan and Li, Yandong and Jaakkola, Tommi and Jia, Xuhui and others},
  journal={arXiv preprint arXiv:2501.09732},
  year={2025}
}

@inproceedings{rombach2022high,
  title={High-resolution image synthesis with latent diffusion models},
  author={Rombach, Robin and Blattmann, Andreas and Lorenz, Dominik and Esser, Patrick and Ommer, Bj{\"o}rn},
  booktitle={Proceedings of the IEEE/CVF conference on computer vision and pattern recognition},
  pages={10684--10695},
  year={2022}
}

@article{ho2020denoising,
  title={Denoising diffusion probabilistic models},
  author={Ho, Jonathan and Jain, Ajay and Abbeel, Pieter},
  journal={Advances in neural information processing systems},
  volume={33},
  pages={6840--6851},
  year={2020}
}

@article{li2024autoregressive,
  title={Autoregressive image generation without vector quantization},
  author={Li, Tianhong and Tian, Yonglong and Li, He and Deng, Mingyang and He, Kaiming},
  journal={Advances in Neural Information Processing Systems},
  volume={37},
  pages={56424--56445},
  year={2024}
}

@article{tian2024visual,
  title={Visual autoregressive modeling: Scalable image generation via next-scale prediction},
  author={Tian, Keyu and Jiang, Yi and Yuan, Zehuan and Peng, Bingyue and Wang, Liwei},
  journal={Advances in neural information processing systems},
  volume={37},
  pages={84839--84865},
  year={2024}
}

@article{ma2025token,
  title={Token-shuffle: Towards high-resolution image generation with autoregressive models},
  author={Ma, Xu and Sun, Peize and Ma, Haoyu and Tang, Hao and Ma, Chih-Yao and Wang, Jialiang and Li, Kunpeng and Dai, Xiaoliang and Shi, Yujun and Ju, Xuan and others},
  journal={arXiv preprint arXiv:2504.17789},
  year={2025}
}

@article{sun2024autoregressive,
  title={Autoregressive model beats diffusion: Llama for scalable image generation},
  author={Sun, Peize and Jiang, Yi and Chen, Shoufa and Zhang, Shilong and Peng, Bingyue and Luo, Ping and Yuan, Zehuan},
  journal={arXiv preprint arXiv:2406.06525},
  year={2024}
}

@article{ghosh2023geneval,
  title={Geneval: An object-focused framework for evaluating text-to-image alignment},
  author={Ghosh, Dhruba and Hajishirzi, Hannaneh and Schmidt, Ludwig},
  journal={Advances in Neural Information Processing Systems},
  volume={36},
  pages={52132--52152},
  year={2023}
}

@article{wang2025promptenhancer,
  title={Promptenhancer: A simple approach to enhance text-to-image models via chain-of-thought prompt rewriting},
  author={Wang, Linqing and Xing, Ximing and Cheng, Yiji and Zhao, Zhiyuan and Li, Donghao and Hang, Tiankai and Tao, Jiale and Wang, Qixun and Li, Ruihuang and Chen, Comi and others},
  journal={arXiv preprint arXiv:2509.04545},
  year={2025}
}

@article{cai2025z,
  title={Z-Image: An Efficient Image Generation Foundation Model with Single-Stream Diffusion Transformer},
  author={Cai, Huanqia and Cao, Sihan and Du, Ruoyi and Gao, Peng and Hoi, Steven and Hou, Zhaohui and Huang, Shijie and Jiang, Dengyang and Jin, Xin and Li, Liangchen and others},
  journal={arXiv preprint arXiv:2511.22699},
  year={2025}
}

@article{huang2023t2i,
  title={T2i-compbench: A comprehensive benchmark for open-world compositional text-to-image generation},
  author={Huang, Kaiyi and Sun, Kaiyue and Xie, Enze and Li, Zhenguo and Liu, Xihui},
  journal={Advances in Neural Information Processing Systems},
  volume={36},
  pages={78723--78747},
  year={2023}
}

@article{wu2023human,
  title={Human preference score v2: A solid benchmark for evaluating human preferences of text-to-image synthesis},
  author={Wu, Xiaoshi and Hao, Yiming and Sun, Keqiang and Chen, Yixiong and Zhu, Feng and Zhao, Rui and Li, Hongsheng},
  journal={arXiv preprint arXiv:2306.09341},
  year={2023}
}

@article{podell2023sdxl,
  title={Sdxl: Improving latent diffusion models for high-resolution image synthesis},
  author={Podell, Dustin and English, Zion and Lacey, Kyle and Blattmann, Andreas and Dockhorn, Tim and M{\"u}ller, Jonas and Penna, Joe and Rombach, Robin},
  journal={arXiv preprint arXiv:2307.01952},
  year={2023}
}

@article{bai2025qwen2,
  title={Qwen2. 5-vl technical report},
  author={Bai, Shuai and Chen, Keqin and Liu, Xuejing and Wang, Jialin and Ge, Wenbin and Song, Sibo and Dang, Kai and Wang, Peng and Wang, Shijie and Tang, Jun and others},
  journal={arXiv preprint arXiv:2502.13923},
  year={2025}
}

@inproceedings{ma2025hpsv3,
  title={Hpsv3: Towards wide-spectrum human preference score},
  author={Ma, Yuhang and Wu, Xiaoshi and Sun, Keqiang and Li, Hongsheng},
  booktitle={Proceedings of the IEEE/CVF International Conference on Computer Vision},
  pages={15086--15095},
  year={2025}
}

@misc{labs2025flux1kontextflowmatching,
      title={FLUX.1 Kontext: Flow Matching for In-Context Image Generation and Editing in Latent Space},
      author={Black Forest Labs and Stephen Batifol and Andreas Blattmann and Frederic Boesel and Saksham Consul and Cyril Diagne and Tim Dockhorn and Jack English and Zion English and Patrick Esser and Sumith Kulal and Kyle Lacey and Yam Levi and Cheng Li and Dominik Lorenz and Jonas Müller and Dustin Podell and Robin Rombach and Harry Saini and Axel Sauer and Luke Smith},
      year={2025},
      eprint={2506.15742},
      archivePrefix={arXiv},
      primaryClass={cs.GR},
      url={https://arxiv.org/abs/2506.15742},
}

@misc{flux2024,
    author={Black Forest Labs},
    title={FLUX},
    year={2024},
    howpublished={\url{https://github.com/black-forest-labs/flux}},
}

@article{chen2025blip3,
  title={Blip3-o: A family of fully open unified multimodal models-architecture, training and dataset},
  author={Chen, Jiuhai and Xu, Zhiyang and Pan, Xichen and Hu, Yushi and Qin, Can and Goldstein, Tom and Huang, Lifu and Zhou, Tianyi and Xie, Saining and Savarese, Silvio and others},
  journal={arXiv preprint arXiv:2505.09568},
  year={2025}
}

@misc{jackyhate2024texttoimage2m,
  author       = {jackyhate},
  title        = {text-to-image-2M},
  year         = {2024},
  publisher    = {Hugging Face},
  howpublished = {\url{https://huggingface.co/datasets/jackyhate/text-to-image-2M}},
  doi          = {10.57967/hf/3066}
}

@article{kirstain2023pick,
  title={Pick-a-pic: An open dataset of user preferences for text-to-image generation},
  author={Kirstain, Yuval and Polyak, Adam and Singer, Uriel and Matiana, Shahbuland and Penna, Joe and Levy, Omer},
  journal={Advances in neural information processing systems},
  volume={36},
  pages={36652--36663},
  year={2023}
}

@article{gong2025seedream,
  title={Seedream 2.0: A native chinese-english bilingual image generation foundation model},
  author={Gong, Lixue and Hou, Xiaoxia and Li, Fanshi and Li, Liang and Lian, Xiaochen and Liu, Fei and Liu, Liyang and Liu, Wei and Lu, Wei and Shi, Yichun and others},
  journal={arXiv preprint arXiv:2503.07703},
  year={2025}
}

@article{liu2025flow,
  title={Flow-grpo: Training flow matching models via online rl},
  author={Liu, Jie and Liu, Gongye and Liang, Jiajun and Li, Yangguang and Liu, Jiaheng and Wang, Xintao and Wan, Pengfei and Zhang, Di and Ouyang, Wanli},
  journal={arXiv preprint arXiv:2505.05470},
  year={2025}
}

@article{liu2025understanding,
  title={Understanding r1-zero-like training: A critical perspective},
  author={Liu, Zichen and Chen, Changyu and Li, Wenjun and Qi, Penghui and Pang, Tianyu and Du, Chao and Lee, Wee Sun and Lin, Min},
  journal={arXiv preprint arXiv:2503.20783},
  year={2025}
}

@article{seedream2025seedream,
  title={Seedream 4.0: Toward next-generation multimodal image generation},
  author={Seedream, Team and Chen, Yunpeng and Gao, Yu and Gong, Lixue and Guo, Meng and Guo, Qiushan and Guo, Zhiyao and Hou, Xiaoxia and Huang, Weilin and Huang, Yixuan and others},
  journal={arXiv preprint arXiv:2509.20427},
  year={2025}
}

@article{sun2023journeydb,
  title={Journeydb: A benchmark for generative image understanding},
  author={Sun, Keqiang and Pan, Junting and Ge, Yuying and Li, Hao and Duan, Haodong and Wu, Xiaoshi and Zhang, Renrui and Zhou, Aojun and Qin, Zipeng and Wang, Yi and others},
  journal={Advances in neural information processing systems},
  volume={36},
  pages={49659--49678},
  year={2023}
}

@article{Qwen3-VL,
      title={Qwen3-VL Technical Report}, 
      author={Shuai Bai and Yuxuan Cai and Ruizhe Chen and Keqin Chen and Xionghui Chen and Zesen Cheng and Lianghao Deng and Wei Ding and Chang Gao and Chunjiang Ge and Wenbin Ge and Zhifang Guo and Qidong Huang and Jie Huang and Fei Huang and Binyuan Hui and Shutong Jiang and Zhaohai Li and Mingsheng Li and Mei Li and Kaixin Li and Zicheng Lin and Junyang Lin and Xuejing Liu and Jiawei Liu and Chenglong Liu and Yang Liu and Dayiheng Liu and Shixuan Liu and Dunjie Lu and Ruilin Luo and Chenxu Lv and Rui Men and Lingchen Meng and Xuancheng Ren and Xingzhang Ren and Sibo Song and Yuchong Sun and Jun Tang and Jianhong Tu and Jianqiang Wan and Peng Wang and Pengfei Wang and Qiuyue Wang and Yuxuan Wang and Tianbao Xie and Yiheng Xu and Haiyang Xu and Jin Xu and Zhibo Yang and Mingkun Yang and Jianxin Yang and An Yang and Bowen Yu and Fei Zhang and Hang Zhang and Xi Zhang and Bo Zheng and Humen Zhong and Jingren Zhou and Fan Zhou and Jing Zhou and Yuanzhi Zhu and Ke Zhu},
	  journal={arXiv preprint arXiv:2511.21631},
      year={2025}
}

@article{inan2023llama,
  title={Llama guard: Llm-based input-output safeguard for human-ai conversations},
  author={Inan, Hakan and Upasani, Kartikeya and Chi, Jianfeng and Rungta, Rashi and Iyer, Krithika and Mao, Yuning and Tontchev, Michael and Hu, Qing and Fuller, Brian and Testuggine, Davide and others},
  journal={arXiv preprint arXiv:2312.06674},
  year={2023}
}

@article{reimers2019sentence,
  title={Sentence-bert: Sentence embeddings using siamese bert-networks},
  author={Reimers, Nils and Gurevych, Iryna},
  journal={arXiv preprint arXiv:1908.10084},
  year={2019}
}

@article{wang2020minilm,
  title={Minilm: Deep self-attention distillation for task-agnostic compression of pre-trained transformers},
  author={Wang, Wenhui and Wei, Furu and Dong, Li and Bao, Hangbo and Yang, Nan and Zhou, Ming},
  journal={Advances in neural information processing systems},
  volume={33},
  pages={5776--5788},
  year={2020}
}

@article{cao2025hunyuanimage,
  title={Hunyuanimage 3.0 technical report},
  author={Cao, Siyu and Chen, Hangting and Chen, Peng and Cheng, Yiji and Cui, Yutao and Deng, Xinchi and Dong, Ying and Gong, Kipper and Gu, Tianpeng and Gu, Xiusen and others},
  journal={arXiv preprint arXiv:2509.23951},
  year={2025}
}

@article{grattafiori2024llama,
  title={The llama 3 herd of models},
  author={Grattafiori, Aaron and Dubey, Abhimanyu and Jauhri, Abhinav and Pandey, Abhinav and Kadian, Abhishek and Al-Dahle, Ahmad and Letman, Aiesha and Mathur, Akhil and Schelten, Alan and Vaughan, Alex and others},
  journal={arXiv preprint arXiv:2407.21783},
  year={2024}
}

@article{comanici2025gemini,
  title={Gemini 2.5: Pushing the frontier with advanced reasoning, multimodality, long context, and next generation agentic capabilities},
  author={Comanici, Gheorghe and Bieber, Eric and Schaekermann, Mike and Pasupat, Ice and Sachdeva, Noveen and Dhillon, Inderjit and Blistein, Marcel and Ram, Ori and Zhang, Dan and Rosen, Evan and others},
  journal={arXiv preprint arXiv:2507.06261},
  year={2025}
}

@misc{flux-2-2025,
    author={Black Forest Labs},
    title={{FLUX.2: Frontier Visual Intelligence}},
    year={2025},
    howpublished={\url{https://bfl.ai/blog/flux-2}},
}

@article{gao2025seedream,
  title={Seedream 3.0 technical report},
  author={Gao, Yu and Gong, Lixue and Guo, Qiushan and Hou, Xiaoxia and Lai, Zhichao and Li, Fanshi and Li, Liang and Lian, Xiaochen and Liao, Chao and Liu, Liyang and others},
  journal={arXiv preprint arXiv:2504.11346},
  year={2025}
}

@article{kuznetsova2020open,
  title={The open images dataset v4: Unified image classification, object detection, and visual relationship detection at scale},
  author={Kuznetsova, Alina and Rom, Hassan and Alldrin, Neil and Uijlings, Jasper and Krasin, Ivan and Pont-Tuset, Jordi and Kamali, Shahab and Popov, Stefan and Malloci, Matteo and Kolesnikov, Alexander and others},
  journal={International journal of computer vision},
  volume={128},
  number={7},
  pages={1956--1981},
  year={2020},
  publisher={Springer}
}

@inproceedings{changpinyo2021conceptual,
  title={Conceptual 12m: Pushing web-scale image-text pre-training to recognize long-tail visual concepts},
  author={Changpinyo, Soravit and Sharma, Piyush and Ding, Nan and Soricut, Radu},
  booktitle={Proceedings of the IEEE/CVF conference on computer vision and pattern recognition},
  pages={3558--3568},
  year={2021}
}

@article{schuhmann2022laion,
  title={Laion-5b: An open large-scale dataset for training next generation image-text models},
  author={Schuhmann, Christoph and Beaumont, Romain and Vencu, Richard and Gordon, Cade and Wightman, Ross and Cherti, Mehdi and Coombes, Theo and Katta, Aarush and Mullis, Clayton and Wortsman, Mitchell and others},
  journal={Advances in neural information processing systems},
  volume={35},
  pages={25278--25294},
  year={2022}
}

@inproceedings{deng2025openvlthinker,
  title={Openvlthinker: Complex vision-language reasoning via iterative sft-rl cycles},
  author={Deng, Yihe and Bansal, Hritik and Yin, Fan and Peng, Nanyun and Wang, Wei and Chang, Kai-Wei},
  booktitle={The Thirty-ninth Annual Conference on Neural Information Processing Systems},
  year={2025}
}

@article{wang2025ovis,
  title={Ovis-Image Technical Report},
  author={Wang, Guo-Hua and Cao, Liangfu and Cui, Tianyu and Fu, Minghao and Chen, Xiaohao and Zhan, Pengxin and Zhao, Jianshan and Li, Lan and Fu, Bowen and Liu, Jiaqi and others},
  journal={arXiv preprint arXiv:2511.22982},
  year={2025}
}

@article{wu2025rewarddance,
  title={Rewarddance: Reward scaling in visual generation},
  author={Wu, Jie and Gao, Yu and Ye, Zilyu and Li, Ming and Li, Liang and Guo, Hanzhong and Liu, Jie and Xue, Zeyue and Hou, Xiaoxia and Liu, Wei and others},
  journal={arXiv preprint arXiv:2509.08826},
  year={2025}
}

@article{wu2025qwen,
  title={Qwen-image technical report},
  author={Wu, Chenfei and Li, Jiahao and Zhou, Jingren and Lin, Junyang and Gao, Kaiyuan and Yan, Kun and Yin, Sheng-ming and Bai, Shuai and Xu, Xiao and Chen, Yilei and others},
  journal={arXiv preprint arXiv:2508.02324},
  year={2025}
}
\bibliographystyle{icml2026}

\newpage
\newpage
\appendix
\section{Detailed Distribution analysis}
\label{sec:detailed_distribution_analysis}
\begin{figure}[!h]
    \centering
    \begin{subfigure}[b]{1\linewidth}
        \centering
        \includegraphics[width=\linewidth]{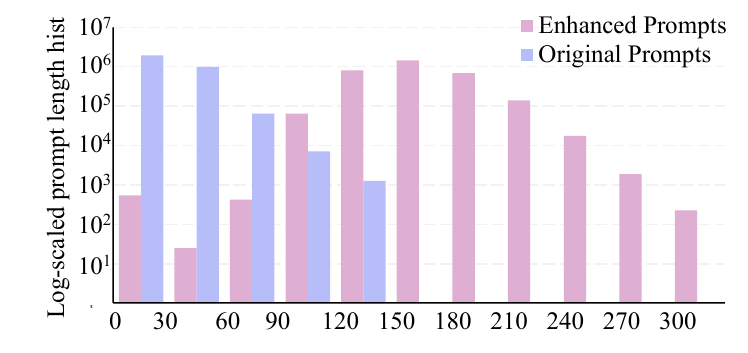}
        \caption{synthetic sets prompt length distribution.} 
        \label{fig:prompt_length_synthetic}
    \end{subfigure}
    \hfill 
    \begin{subfigure}[b]{1\linewidth}
        \centering
        \includegraphics[width=\linewidth]{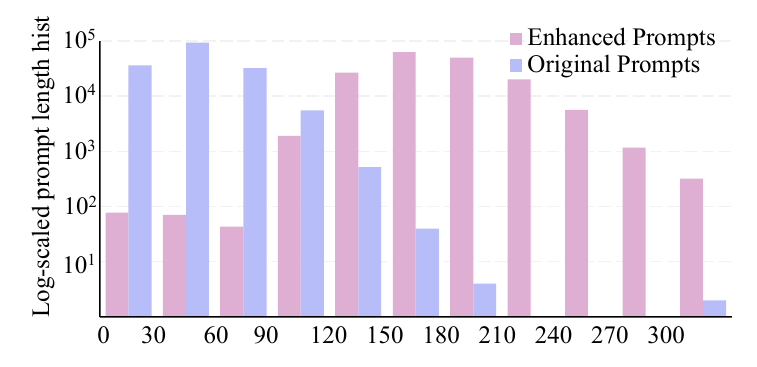}
        \caption{curated real-image set prompt length distribution.} 
        \label{fig:prompt_length_curated}
    \end{subfigure}
    \hfill 
    \caption{Prompt length distribution analysis.}
    \label{fig:prompt_length}
\end{figure}

In this section, we provide a more detailed analysis of our Fine-T2I dataset. Table~\ref{tab:statistics_style} summarizes style statistics for our synthetic sets. To further characterize the data, Fig.~\ref{fig:prompt_length} shows the prompt-length distribution. Original prompts are generally short (typically under 50 words), whereas enhanced prompts are substantially longer and span a wider range of lengths. We observe the same trend in both the synthetic sets and the curated real-image set. Finally, we report detailed statistics for categories and tasks in Table~\ref{tab:statistics_category} and Table~\ref{tab:statistics_task}, respectively. For transparency, we also report our pre-defined resolutions and aspect ratios for creating synthetic images in Table~\ref{tab:predefined_resolutions}.

To ensure diversity in prompt structure and template, we design five distinct instructions (which is added to the system prompts) that guide LLaMA3 during prompt generation. These templates are listed as follows:
\begin{itemize}
    \item simple structure starts like: a photo of xxx, an image of xxx, a picture of xxxx, etc.
    \item like user is asking for a request, for examples, please help generate xxx, Could you please help provide an image xxxx, generate an image of, etc.
    \item one or multiple sentences.
    \item combinations of sentence(s) and words (e.g., a sentence of multiple sentence, + 8K resolution, photorealistic, natural, etc.), etc.
    \item simple combination of of words (e.g., dog, flying, over sea, etc.).
\end{itemize}

\begin{table}[!h]
\caption{Detailed statistics for styles in our Fine-T2I four synthetic subsets. ``PE" indicates enhanced prompt, ``PO" indicates original prompt, ``AR" means random pre-defined aspect ratios, and ``AS" means square aspect ratio. This table shows the detailed sample numbers belonging to each pre-defined style.}
\label{tab:statistics_style}
\resizebox{1\linewidth}{!}{
\begin{tabular}{llllll}
\toprule
\textbf{Styles}                              & \textbf{PE-AR}  & \textbf{PE-AS}  & \textbf{PO-AR}  & \textbf{PO-AS}  & \textbf{Total}   \\
\midrule
3D \& CGI                           & 60309  & 58141  & 62090  & 47831  & 228371  \\
Abstract \& Artistic                & 143509 & 138761 & 139430 & 109619 & 531319  \\
Anime                               & 189551 & 183866 & 189392 & 143096 & 705905  \\
Cartoon \& Illustration             & 134538 & 132375 & 140222 & 112841 & 519976  \\
Cultural \& Folk Art                & 141827 & 128671 & 161804 & 123120 & 555422  \\
Futuristic \& Sci-Fi                & 92631  & 92110  & 86677  & 59929  & 331347  \\
Graphic\& Digital                   & 210362 & 200629 & 221038 & 175241 & 807270  \\
Others                              & 36275  & 34006  & 37378  & 29387  & 137046  \\
Traditional/modern Art           & 155099 & 149882 & 159576 & 125033 & 589590  \\
Vintage \& Retro                    & 124268 & 114807 & 151822 & 115733 & 506630 \\
General \& Photorealistic           & 327223 & 305005 & 337069 & 263520 & 1232817 \\
\bottomrule
\end{tabular}
}
\end{table}

\begin{table}[]
\caption{Detailed statistics for categories in our Fine-T2I four synthetic subsets.}
\label{tab:statistics_category}
\resizebox{1\linewidth}{!}{
\begin{tabular}{lllllll}                                \\
\toprule
\multicolumn{2}{c}{\textbf{Category}}                            & \textbf{PE-AR}  & \textbf{PE-AS}  & \textbf{PO-AR}  & \textbf{PO-AS}  & \textbf{Total}   \\
\midrule
\multirow{8}{*}{\textbf{Design}}         & Arts                     & 59722  & 56909  & 67021  & 51893  & 235545 \\
                                & Cartoon                  & 21975  & 22994  & 25973  & 20517  & 91459  \\
                                & Creative                 & 3734   & 3593   & 4245   & 3267   & 14839  \\
                                & Logos                    & 3434   & 3616   & 3240   & 2489   & 12779  \\
                                & Posters                  & 28819  & 27444  & 28142  & 21159  & 105564 \\
                                & Rare cases               & 4963   & 4736   & 5654   & 4050   & 19403  \\
                                & Slides                   & 19475  & 18751  & 17708  & 13326  & 69260  \\
                                & UI/UX                    & 28266  & 27310  & 28417  & 20424  & 104417 \\
\midrule
\multirow{9}{*}{\textbf{Nature}}         & Animals                  & 59710  & 57187  & 66406  & 51122  & 234425 \\
                                & Cityscape                & 64063  & 58235  & 71190  & 52812  & 246300 \\
                                & Creative                 & 4540   & 4336   & 4703   & 3694   & 17273  \\
                                & Food                     & 61647  & 56695  & 77209  & 61570  & 257121 \\
                                & Indoor                   & 47106  & 43443  & 57773  & 45450  & 193772 \\
                                & Landscape                & 35308  & 33341  & 35869  & 28995  & 133513 \\
                                & Objects / Entities       & 122078 & 115692 & 131568 & 102185 & 471523 \\
                                & Plants                   & 34259  & 32686  & 36440  & 29212  & 132597 \\
                                & Rare cases               & 6330   & 5869   & 6187   & 4674   & 23060  \\
                                \midrule
\multirow{7}{*}{\textbf{People}}         & Activities               & 149561 & 136908 & 174234 & 136109 & 596812 \\
                                & Age                      & 61683  & 57572  & 67566  & 54927  & 241748 \\
                                & Creative                 & 11868  & 11057  & 13494  & 10478  & 46897  \\
                                & Emotions                 & 98799  & 93809  & 101281 & 82550  & 376439 \\
                                & Portrait                 & 219778 & 209840 & 242322 & 197778 & 869718 \\
                                & Rare cases               & 10186  & 9289   & 8822   & 6290   & 34587  \\
                                & Sports                   & 42077  & 38626  & 47790  & 33910  & 162403 \\
                                \midrule
\textbf{Rare }                           & Rare cases               & 104062 & 100557 & 105787 & 76979  & 387385 \\
\midrule
\multirow{7}{*}{\textbf{Text rendering}} & Creative                 & 5466   & 5425   & 4755   & 3598   & 19244  \\
                                & Handwriting              & 13841  & 13614  & 11586  & 8692   & 47733  \\
                                & Long text                & 74115  & 70762  & 43075  & 29381  & 217333 \\
                                & Rare cases               & 3938   & 3770   & 3172   & 2209   & 13089  \\
                                & Short text               & 111131 & 110970 & 101908 & 77247  & 401256 \\
                                & Stylized text & 23797  & 24374  & 20920  & 15560  & 84651  \\
                                & Text in scene            & 79861  & 78843  & 72041  & 52803  & 283548 \\
\bottomrule
\end{tabular}
}
\end{table}

\begin{table}[!h]
\caption{Detailed statistics for tasks in our Fine-T2I four synthetic subsets. For general generation, over half of the prompts are not assigned a particular task.}
\label{tab:statistics_task}
\resizebox{1\linewidth}{!}{
\begin{tabular}{llllll}
\toprule
\textbf{Tasks}                              & \textbf{PE-AR}  & \textbf{PE-AS}  & \textbf{PO-AR}  & \textbf{PO-AS}  & \textbf{Total}   \\
\midrule
\textbf{Colors}              & 84885  & 81386  & 89151   & 70316  & 325738  \\
\textbf{Counting}            & 83826  & 77285  & 76807   & 56341  & 294259  \\
\textbf{Position}            & 101398 & 96048  & 103106  & 80280  & 380832  \\
\textbf{Reasoning}           & 127420 & 118657 & 136928  & 106034 & 489039  \\
\midrule
\textbf{Colors + Counting}    & 29754  & 27932  & 25489   & 18589  & 101764  \\
\textbf{Colors + Position}    & 40304  & 38229  & 40570   & 31972  & 151075  \\
\textbf{Colors + Reasoning}   & 46128  & 43543  & 48277   & 37638  & 175586  \\
\textbf{Counting + Position}  & 37795  & 35262  & 33496   & 24721  & 131274  \\
\textbf{Counting + Reasoning} & 38366  & 35129  & 34176   & 24983  & 132654  \\
\textbf{Position + Reasoning} & 49698  & 46546  & 50872   & 39317  & 186433 \\
\midrule
\textbf{Not Specified}                & 976018 & 938236 & 1047626 & 815159 & 3777039 \\
\bottomrule
\end{tabular}
}
\end{table}

\begin{table}[!h]
\centering
\caption{Pre-defined resolutions and aspect ratios for generating synthetic images.}
\label{tab:predefined_resolutions}
\resizebox{0.7\linewidth}{!}{
\begin{tabular}{lcc}
\toprule
                            & \textbf{Aspect ratio} & \textbf{Resolution}       \\
                            \midrule
\multirow{6}{*}{\textbf{Square}}     & 1:1          & {[}512, 512{]}   \\
                            & 1:1          & {[}768, 768{]}   \\
                            & 1:1          & {[}1024, 1024{]} \\
                            & 1:1          & {[}1536, 1536{]} \\
                            & 1:1          & {[}2048, 2048{]} \\
                            & 1:1          & {[}2560, 2560{]} \\
                            \midrule
\multirow{10}{*}{\textbf{Landscape}} & 4:3          & {[}2048, 1536{]} \\
                            & 4:3          & {[}1024, 768{]}  \\
                            & 4:3          & {[}1600, 1200{]} \\
                            & 3:2          & {[}1152, 768{]}  \\
                            & 3:2          & {[}1536, 1024{]} \\
                            & 16:9         & {[}1280, 720{]}  \\
                            & 16:9         & {[}2048, 1152{]} \\
                            & 16:9         & {[}2560, 1440{]} \\
                            & 5:4          & {[}1280, 1024{]} \\
                            & 5:4          & {[}1920, 1536{]} \\
                            \midrule
\multirow{10}{*}{\textbf{Portrait}}  & 3:4          & {[}768, 1024{]}  \\
                            & 3:4          & {[}1200, 1600{]} \\
                            & 3:4          & {[}1536, 2048{]} \\
                            & 2:3          & {[}768, 1152{]}  \\
                            & 2:3          & {[}1024, 1536{]} \\
                            & 9:16         & {[}720, 1280{]}  \\
                            & 9:16         & {[}1152, 2048{]} \\
                            & 9:16         & {[}1440, 2560{]} \\
                            & 4:5          & {[}1024, 1280{]} \\
                            & 4:5          & {[}1536, 1920{]} \\
                            \bottomrule
\end{tabular}
}
\end{table}

\section{Problems, Ambiguity, Expectation}
\label{sec:problems}
\paragraph{Diversity of generated Prompt} 
During prompt synthesis, we observed severe redundancy. Prompts generated across different batches and even across different GPUs were often highly similar, and in many cases even identical. Assigning different random seeds for each GPU did not help the problem a lot. As a mitigation, we increased the LLM sampling temperature to encourage stochasticity. However, this also raised the variance of outputs and occasionally produced malformed generations that were not valid prompts (which we filtered out later). Instead, conditioning on predefined attributes such as style and category further alleviated the issue, but substantial duplication persisted and we did not find a principled way to eliminate it. Consequently, the deduplication stage alone removed nearly 90\% of the generated candidates, leading to significant wasted computation.

\paragraph{Attributes alignment with prompt} 
Although we apply attribute-based filtering in Sec.~\ref{sec:prompt_filtering}, we do not guarantee the attributes (category, style, and tasks) can be perfectly aligned with the resulting prompt. In practice, some attribute combinations are internally inconsistent or overly constrained. For example, we may request an extremely concise prompt (\textit{e.g.}, less than 10 words) while simultaneously requiring a long textual rendering description, or impose strong multi-factor constraints such as UI/UX design + cultural style + counting and colors tasks. Such contradictory or complex conditions can prevent the attributes from being faithfully expressed in the generated prompt. We therefore recommend treating the attributes as soft metadata when using the dataset. They are useful for reference and analysis but may not be accurate, while we ensure the text–image pairs remain reliable for training and evaluation. If stricter attribute adherence is required, one can apply an additional LLM-based reasoning filter (as in our later pipeline stages in Sec.~\ref{sec:text_image_pair_filtering}), but this typically results in substantially more aggressive filtering.

\begin{figure}[!h]
    \centering
    \includegraphics[width=1\linewidth]{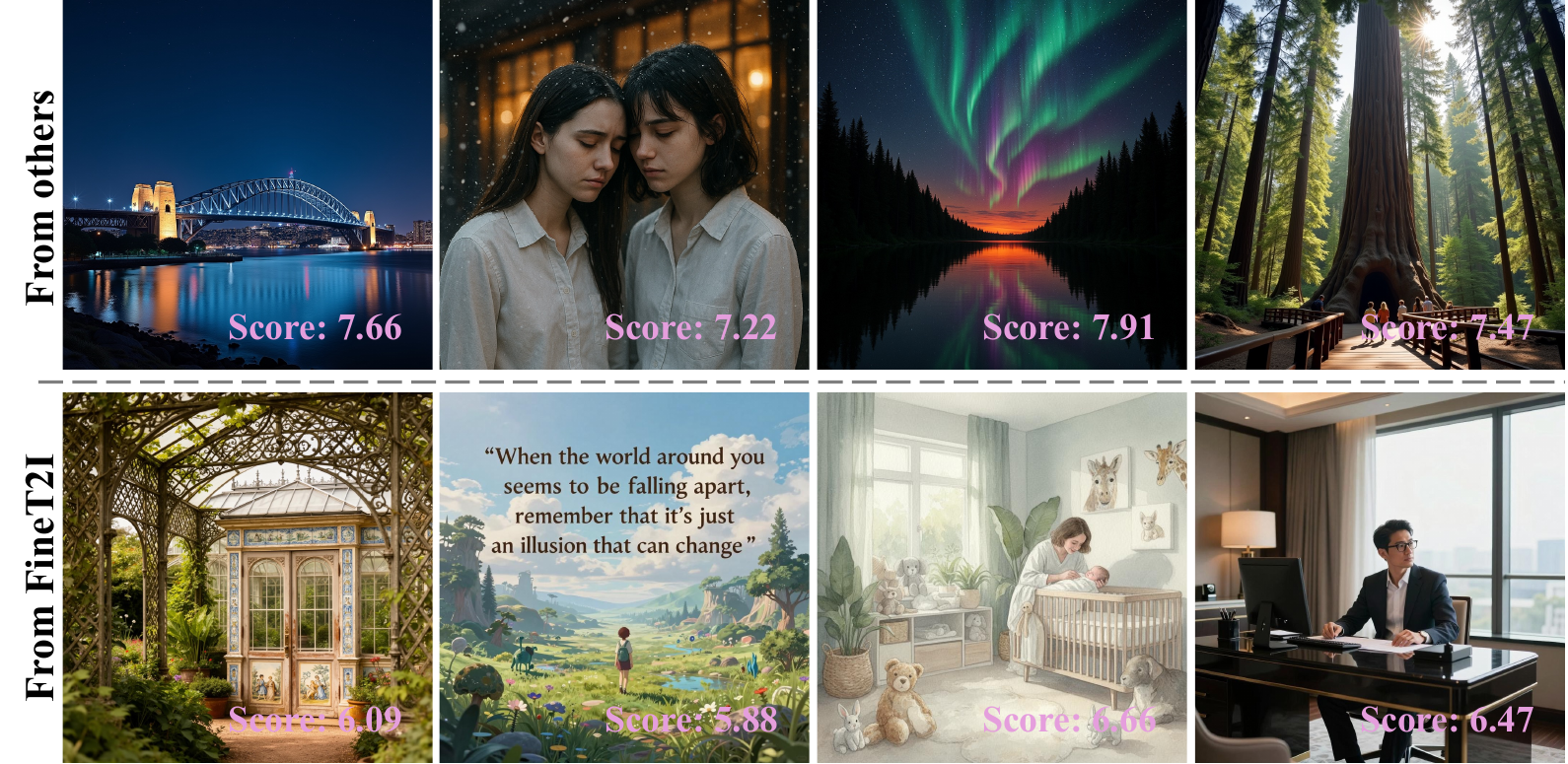}
    \caption{Aesthetics scores may be ambiguous and may not be aligned with human preference. Indeed, aesthetics always change with the development and can be biased. Please zoom in to see details.}
    \label{fig:aesthetics_problem}
\end{figure}

\paragraph{Ambiguity of aesthetics and human preference}
What people call a “high-quality” image in text-to-image generation has been changing quickly. A few years ago, highly stylized, visually appearing generations were often considered as higher quality. Instead, people now prefer realistic, coherent, and photograph-like generations. This makes “aesthetics” an inherently ambiguous target for dataset collection, and a single automatic aesthetic score (or even a combination of multiple scores) can disagree with what humans actually prefer. Fig.~\ref{fig:aesthetics_problem} shows a representative example. The top row enjoys a higher aesthetic score, yet people may believe the bottom row is better because it looks more natural. Given this mismatch, we do not treat any aesthetic metric as ground truth. Instead, we follow a pragmatic criterion aligned with current preferences, using the quality signals embodied by strong image generators when generating our synthetic sets, and we trust the taste of professional photographers when collecting our curated real-image set.

\paragraph{Expectation for better image evaluation}
Building an extremely high-quality fine-tuning dataset required many, sometimes even cumbersome filtering steps.
We rely on heavy filtering because there is still no strong, robust, widely accepted indicator for assessing the quality of an image or an image–text pair.
We expect that a strong and reliable evaluation metric would substantially simplify this process, and we believe such a metric may benefit from the reasoning capabilities of VLMs.

\begin{figure}[!t]
    \centering
    \includegraphics[width=1\linewidth]{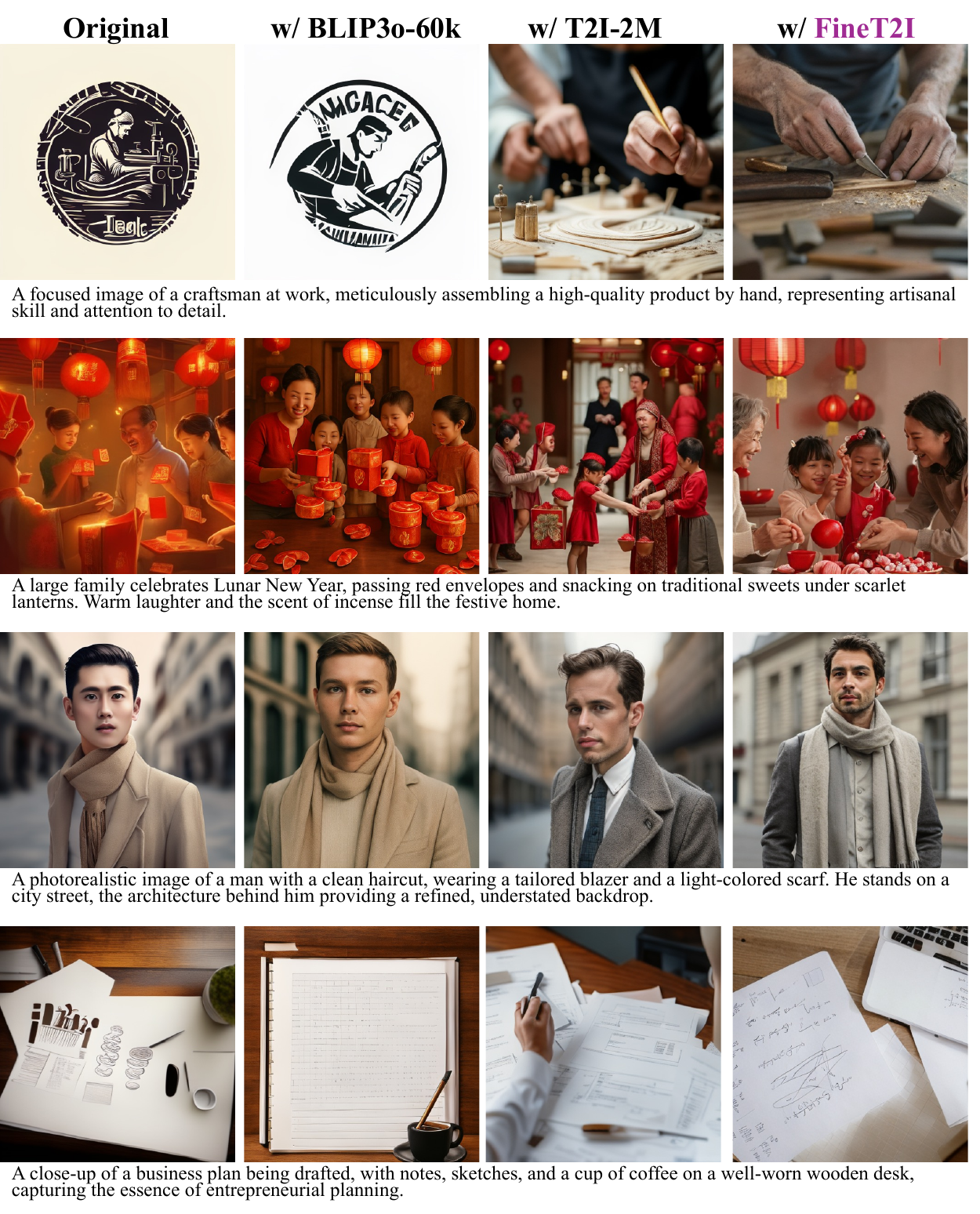}
    \caption{Visual comparison of LlamaGen fine-tuned on different datasets. We randomly pick examples from generated images for visualization.}
    \label{fig:compare_4}
\end{figure}

\begin{figure*}[!h]  
\centering
\begin{minipage}{0.96\textwidth}  
\begin{chatbox}{System Prompt}
You are a Strict Visual Quality Auditor. Your sole task is to perform a binary pass/fail audit on text-image pairs based on absolute prompt alignment and technical flawlessness.

\vspace{2mm}
\textbf{Evaluation Criteria:}

\vspace{2mm}
{\setlength{\leftskip}{2em}
\textbf{I. Semantic \& Text-Image Alignment}

- Object Completeness: Every entity, person, or item required in the prompt must be clearly visible and identifiable. Missing any requested element results in an immediate 'False'.
                
- Quantity Precision: The number of objects in the image must exactly match the count specified in the text.
                - Text Rendering (OCR): Any text, words, or letters requested in the prompt must be rendered with 100 spelling accuracy, correct font style (if specified), and zero character distortion.

- Attribute \& Color Consistency: All specified colors, materials, sizes, and specific properties of objects must be strictly followed.

- Spatial \& Relational Logic: Objects must be in the exact positions described . Actions/interactions between objects must be logical.

- Negative Constraints: If the prompt specifies what not to include, the image must not contain those elements.

\vspace{2mm}
\textbf{II. Technical \& Aesthetic Quality}

                - Anatomical Integrity: Zero tolerance for "AI hallucinations" in biological structures. This includes extra/missing fingers, limbs, distorted faces, unnatural joints, or merged body parts, etc.
                
                - Physical \& Geometric Logic: Objects must follow the laws of physics (unless the prompt says otherwise). No floating objects, impossible perspectives, unrealistic interactions, or "melting" textures where surfaces bleed into each other.
                
                - Image Artifacts: Zero tolerance for unintended blurring, watermarks, signature-like scribbles, garbled characters, distorted characters or text, etc.
                
                - Style \& Medium Fidelity: The image must perfectly embody the requested style (e.g., "CGI," "Oil Painting," "Macro Photography"). If the style is inconsistent across the frame, it is a failure.
                
}

\vspace{2mm}
\textbf{Decision Logic:}

{\setlength{\leftskip}{2em}

            - Output 'True' ONLY if the image is a perfect realization of the prompt with zero technical defects.
            
            - Output 'False' if there is even one minor discrepancy.
            
            - Extra details are permitted only if they enhance the scene without violating the prompt or the laws of physics.
            
}

\vspace{2mm}
\textbf{Constraint:}

{\setlength{\leftskip}{2em}

            - Please be concise but thorough in your thinking.
            
            - You must think step-by-step to verify every detail.
            
            - Output ONLY the word 'True' or 'False' as your final answer.

}

\end{chatbox}
\end{minipage}
\caption{Detailed system prompt used for final text-image pair filtering, for both text-image alignment and aesthetic quality.}
\label{fig:final_filter_system_prompt}
\end{figure*}

\begin{figure*}
    \centering
    \includegraphics[width=1.0\linewidth]{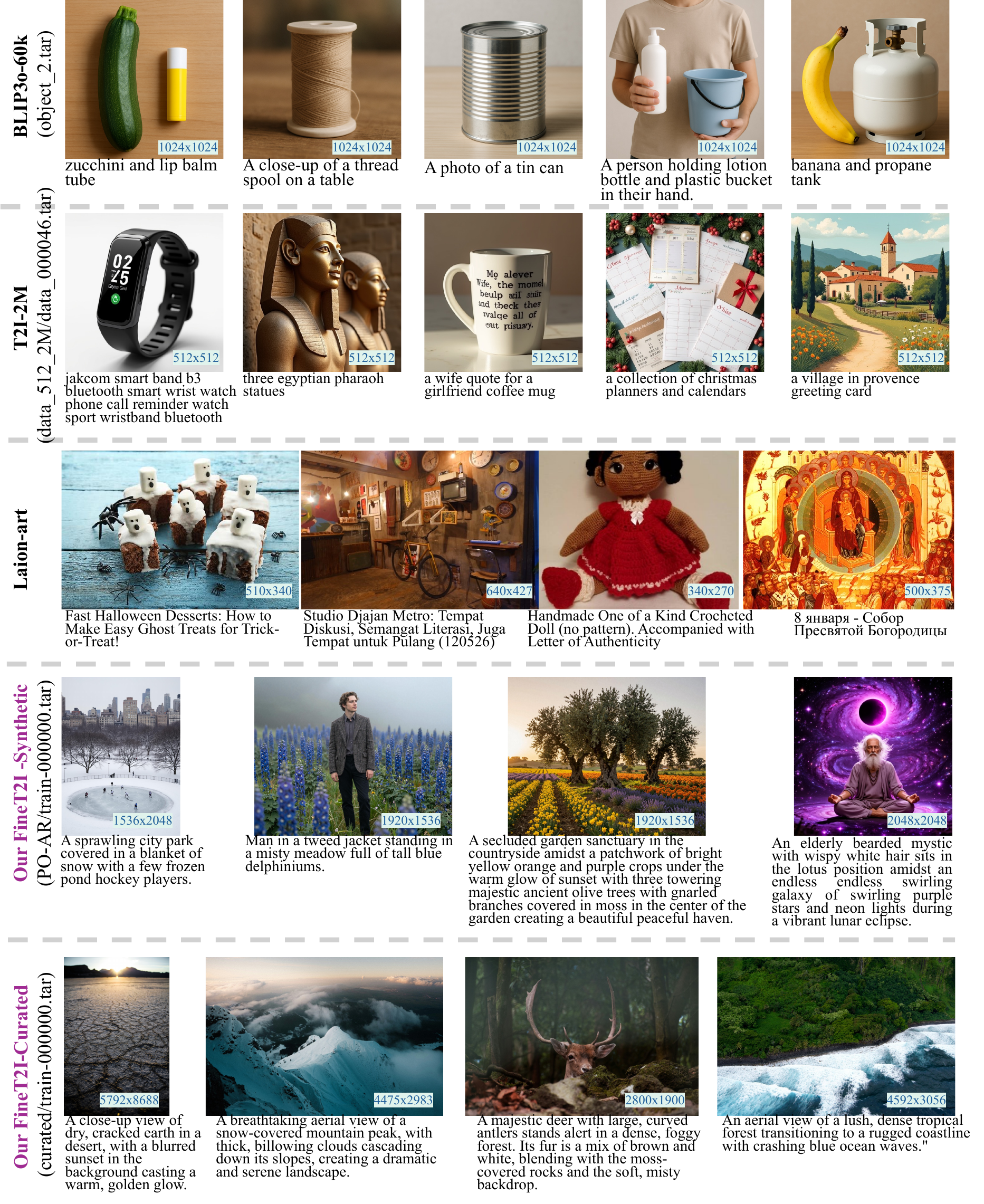}
    \caption{Comparison of text-image pairs from different fine-tuning datasets. We randomly select one tar file from the dataset, and list the first examples as a comparison, without any cherry-picking. The resolution is marked bottom-right. Clearly, our Fine-T2I presents the best alignment, visual quality, and high-resolution, \textit{etc}.}
    \label{fig:compare_dataset}
\end{figure*}

\section{Detailed studies}
We first provide our system prompt for the Qwen-VL thinking model in Fig.~\ref{fig:final_filter_system_prompt},  which ensures high-quality text–image alignment and flawless image generation (explicitly specified in the system prompt).

We also provide samples from different fine-tuning datasets, as shown in Fig.~\ref{fig:compare_dataset}. Instead of cherry-picking specific images, we randomly select a tar file from each dataset and visualize the first several samples. Surprisingly, we observe that some samples from other datasets should not even be included in the fine-tuning set. They are low-resolution, low-quality, and exhibit poor text–image alignment, \textit{etc}., which is far below our expectations for fine-tuning text–image pairs.
This also suggests that both the open community and industry need to put more effort into dataset quality, which is often overlooked.

Next, we show fine-tuning examples on different datasets and present the results in Fig.~\ref{fig:compare_4}. During fine-tuning, we observe that BLIP3o-60k achieves the lowest training loss, but our Fine-T2I clearly delivers the best fine-tuning results. This can be explained by the data characteristics. Images in BLIP3o-60k are generally structurally simple and clear, with limited background and details. This makes it easier for the model to find shortcuts during optimization, but does not indicate better generation performance.

\paragraph{Human Evaluation} We conduct a human study by asking anonymous volunteers to perform the evaluation. Specifically, we suggest that fewer samples should be labeled as “Tie” in the A/B comparison, as shown in Fig.~\ref{fig:human_eval}. For comparing fine-tuning results across different datasets, as shown in Fig.~\ref{fig:human_compare_data}, we explicitly require that one best sample be selected. We report the results averaged over all participating annotators.

\end{document}